\documentclass[twoside,11pt]{article}

\usepackage{blindtext}

\usepackage[preprint]{jmlr2e}

\usepackage{amsmath}
\usepackage{amssymb}
\usepackage{mathtools}
\usepackage{enumitem}

\usepackage{pdflscape}

\usepackage{tikz}
\usetikzlibrary{arrows.meta, positioning}
\usepackage{longtable}
\usepackage{booktabs}
\usepackage{array}
\usetikzlibrary{calc}

\usepackage{microtype}
\usepackage{lastpage}
\jmlrheading{23}{2022}{1-\pageref{LastPage}}{1/21; Revised 5/22}{9/22}{21-0000}{Author One and Author Two}

\ShortHeadings{Biologically Inspired Mechanisms for Grokking}{Leon}
\firstpageno{1}

\begin{document}

\title{Biologically Inspired Mechanisms for Facilitating Grokking in Multilayer Perceptrons}

\author{\name Florin Leon \email florin.leon@academic.tuiasi.ro \\
       \addr Faculty of Automatic Control and Computer Engineering \\
       ``Gheorghe Asachi'' Technical University of Iași \\
        Bd. Mangeron 27, 700050 Iași, Romania
     }


\maketitle

\begin{abstract}
Grokking is a delayed transition from memorization to generalization that is often accompanied by substantial reorganization of internal representations. This paper studies whether biologically inspired mechanisms, many of which are not commonly incorporated into artificial neural networks, can actively promote this transition by regulating hidden-layer computation at the levels of neuronal activity, response, and effective connectivity. We augment a multilayer perceptron with input gating, structural plasticity, gain modulation, threshold modulation, homeostasis, lateral inhibition, and activation decorrelation, and evaluate these mechanisms through systematic ablations on two established grokking benchmarks: sparse parity and noisy XOR classification. The results show that the mechanisms contribute unequally to generalization. Homeostasis provides the strongest and most consistent benefit, while structural sparsification emerges as the second major mechanism. The remaining biologically inspired mechanisms have smaller or less consistent effects in the present experiments. For both problems, the results support the common principle that explicit regulation of neuron utilization and effective connectivity can improve the emergence of generalizable internal computation. These findings motivate broader investigation of biologically inspired activity regulation and adaptive sparsification, including in large language models, where they may accelerate the development of generalizable representations and reduce the optimization time required for robust generalization.
\end{abstract}

\begin{keywords}
grokking, biologically inspired neural mechanisms, hidden representation, structural organization, homeostasis, structural plasticity
\end{keywords}

\section{Introduction}

Grokking denotes a training dynamic in which a neural network can reach near-perfect performance on the training set while test performance remains poor, with strong generalization appearing only after a substantial delay. This temporal gap has been studied most often in small algorithmic tasks, where extended optimization can eventually move the model from memorizing examples toward a more systematic form of generalization that resembles rule learning \citep{power2022grokking}. Further research suggests that this transition is linked to changes in the internal mechanisms of the model, including the emergence of subnetworks that support more generalizable solutions \citep{merrill2023tale}.

This observation points to a broader question about the role of internal structure in delayed generalization. Standard multilayer perceptrons (MLPs), whether trained in the classical backpropagation setting with stochastic gradient descent (SGD) or with more recent first-order optimizers such as AdamW, are designed primarily to minimize prediction error. These training methods impose no direct constraint on the form of the internal organization that the model learns. In particular, they do not ensure modularity, specialization, sparse circuit formation, or any other property that could help the model form robust, reusable internal patterns. Recent work has drawn attention to this issue through the ``fractured entangled representation'' hypothesis \citep{kumar2025questioning}, which argues that conventional objective-driven training can produce representations in which useful regularities are spread across redundant and interfering components.

This question is also relevant beyond small synthetic benchmarks. Large language models (LLMs) also continue to develop and reorganize internal representations over extended training, which makes mechanisms that regulate internal organization relevant at larger scales. If grokking depends in part on the formation of better internal structure, then methods that make that transition faster or more reliable may reduce training cost and improve the emergence of systematic behavior. Therefore, grokking provides a useful setting for studying how architectural biases shape internal representations during learning.

The present paper examines this issue using two benchmarks that have played different roles in the grokking literature. The first is sparse parity \citep{merrill2023tale}, where inputs are binary vectors and the label is determined by the parity of a small hidden subset of coordinates. This task belongs to the class of algorithmic examples that first made grokking visible, and it has since been used to study how a network can move from an early dense solution with poor test performance to a later sparse solution that generalizes well. The second is a noisy XOR-cluster classification \citep{xu2023benign}, where examples are drawn from four Gaussian clusters arranged in a XOR pattern, with a small fraction of labels flipped. In this problem, a neural network can initially fit the training set well but with near-chance test accuracy, and later reach good test performance while still fitting the noisy labels. Sparse parity and noisy XOR reveal two different forms of delayed generalization. In sparse parity, the transition reflects the emergence of a compact subnetwork that captures a hidden dependency among a few input coordinates. In noisy XOR, it reflects a change from sample-level interpolation to features aligned with the nonlinear cluster geometry.

The main difference from most previous works lies in the mechanisms under investigation. Much of the existing grokking literature has focused on standard architectures, optimization effects, and post hoc analysis of the learned solutions. The present study evaluates a set of biologically inspired regulatory mechanisms and examines how they affect grokking behavior. We incorporate these mechanisms into a multilayer perceptron architecture that we call \textit{BioNN}. The mechanisms act within the hidden layers by altering the routing and regulation of activity, the competition among neurons, and the effective connectivity that develops during learning. The goal is to determine whether they can promote a more useful internal organization and whether this organization changes the speed, stability, or form of the transition from memorization to generalization.

The remainder of the paper is organized as follows. Section 2 reviews related work on grokking and brain-inspired neural models. Section 3 introduces the proposed model and describes the biologically inspired mechanisms, including input gating, structural plasticity, gain modulation, threshold modulation, homeostasis, lateral inhibition, and activation decorrelation. Section 4 presents the experimental setup, including the grokking tasks, network structure indicators, and experimental protocol. Section 5 reports the ablation results and discusses the contribution of each mechanism, the organization of the learned representations, implications for generalization and LLMs, and the computational demands of the proposed mechanisms. Section 6 concludes the paper and summarizes the main findings.

The implementation of the model is available at: {\color{blue}https://github.com/florinleon/BioNN}.

\section{Related Work}

\subsection{Grokking}

Existing explanations of grokking can be grouped into a few broad classes. One line of work treats grokking as a consequence of optimization and regularization, where models first reach training interpolation and only later drift toward simpler or lower-norm solutions that generalize better. A second line explains it through delayed feature learning or internal reorganization, where the model gradually develops representations, subnetworks, or circuits aligned with the underlying structure of the task. A third line studies grokking through geometric, statistical or physical perspectives, with properties such as flatness, specialization, phase transitions, or the organization of solution space. Some works move from explanation to intervention and propose training methods designed to accelerate the transition from memorization to generalization.

A survey by \citet{bertolotti2026survey} shows that existing explanations for grokking fall into several broad types. Some accounts present it as a movement among solutions that already fit the training data, where weight decay or related biases gradually favor solutions with smaller norm or lower complexity. Other accounts describe a change in the learning regime, i.e., early training fits the data with features close to initialization, while later training learns features that reflect the structure of the task. A third group of explanations focuses on internal reorganization, such as the emergence of sparse subnetworks, cleaner representations, or circuits that eventually dominate predictions. The survey also shows that these explanations have uneven support for different tasks, architectures, and training settings. 

An empirical study of modular addition is provided by \citet{manir2026systematic}. It shows how depth, architecture, activation function, and regularization shape grokking under comparable training conditions, and argues that grokking is governed less by architecture alone than by the interaction between regularization and the stability of the learning dynamics. Depth does not have a simple monotonic effect, as deeper networks can help or fail depending on whether the learning trajectory remains stable enough to move from memorization to generalization. The comparison between MLPs and Transformers becomes less pronounced once the main training conditions are controlled, which suggests that some apparent architectural differences reflect differences in optimization and regularization. Activation functions affect the timing of grokking, but their effect depends on the regularization regime. The study identifies weight decay as an essential control parameter: when weight decay is too low, the model can fit the training data but remains at low test accuracy; when it is too high, the pressure toward small weights prevents the learning trajectory from forming the task representation that later supports grokking.

The role of weight decay is also analyzed theoretically by \citet{boursier2025theoretical}. Grokking is explained as a two-timescale optimization process with small $L_2$ regularization. Early in training, the gradient from the training loss is much stronger than the contribution of weight decay, so the model behaves almost like an unregularized model and quickly reaches a solution that fits the training data. This early solution can still have a large parameter norm and poor test performance. Once the training loss is close to zero, there are many parameter settings that fit the training data equally well, and the loss provides little preference among them. At this point, weight decay becomes the main source of change. It slowly shifts the parameters within the set of training-fitting solutions to favor points with smaller norm. This explains how test performance can improve long after interpolation; it is because the later phase replaces a high-norm fitting solution with a lower-norm one.

\citet{xu2026grok} study a simpler theoretical case, overparameterized ridge regression. The model has more parameters than training examples, so some parts of the parameter vector can affect new examples without affecting the training loss. Gradient descent first changes the part that is determined by the training data, which quickly lowers the training error. The remaining part is reduced mainly by weight decay, and this reduction is slower. The paper proves that this produces grokking: generalization error stays large after the training error has fallen, then improves later. It also gives bounds on how long this delay can last.

\citet{das2026linear} study a linear classifier trained with logistic loss on separable data. The separating hyperplane passes through the origin, but the trained model also has a learnable bias. The weight vector quickly moves toward the maximum-margin direction. After that, the main remaining change is in the bias. This change depends on the training points closest to the margin (similar to support vectors), and on how those points are distributed between the two classes. The paper shows that this does not usually create grokking on standard test data. Grokking appears when many test examples lie close to the separating hyperplane, because the slow correction of the bias changes their predicted labels only late in training.

\citet{kumar2024lazy} explain grokking as delayed feature learning, or as a transition from lazy training dynamics to a richer feature-learning regime. The delay depends mainly on how quickly the representation changes during training and how well the initial neural tangent kernel is aligned with the target function. When the initial representation is well aligned with the task, the model generalizes early. When the initial representation is poorly aligned, the model first fits the training set using its initial features and generalizes only later, after training changes the representation enough to support the target function. This delayed generalization occurs only in an intermediate data setting. The dataset must be large enough for the model to eventually learn useful features, yet small enough that the model can first reduce training loss without reducing test loss. The paper demonstrates this mechanism with vanilla gradient descent and no weight decay, which shows that norm reduction is in fact not required for grokking. 

\citet{xu2023benign} study a noisy XOR classification problem in which the input distribution has nonlinear structure and a fraction of training labels are flipped. They show that memorization and generalization can emerge at different times in the same model. Very early in training, the network fits all training labels, including the corrupted ones, while clean test performance remains near chance. With further training, test accuracy becomes nearly optimal even though the noisy labels remain memorized. The analysis explains this transition through feature learning, i.e., training gradually builds representations aligned with the true cluster structure, which supports generalization although the memorized noisy labels are still retained.

\citet{merrill2023tale} give a circuit-level account of grokking on the sparse parity task. Before grokking, network behavior is governed by a large, dense subnetwork that fits the training data but does not generalize well. Around the transition, a much smaller set of neurons begins to grow rapidly in norm, while most other neurons decay and become irrelevant. This sparse subnetwork then becomes sufficient to reproduce the full model predictions and appears to implement a parity computation more directly. The paper therefore interprets grokking as a competition between a dense early circuit and a sparse generalizing circuit. 

\citet{rubin2024phase} analyze grokking in two-layer networks trained to equilibrium with Langevin dynamics, and describe hidden activations through their covariance matrix. Grokking occurs when the distribution of hidden weights changes abruptly. At the transition, some hidden neurons move from a task-agnostic state to task-specific states that encode the structure needed for generalization. In a teacher-student model, these neurons align with the teacher direction; in modular addition, they implement Fourier modes. The paper interprets this specialization as a first-order phase transition with Gaussian, mixed, and specialized feature learning. 

\citet{han2025flatness} compare two geometric correlates of generalization: neural collapse and relative flatness. Neural collapse is the tendency of examples from the same class to converge to the same penultimate-layer representation, with class centers and classifier weights settling into a highly symmetric arrangement, whereas relative flatness measures how little the loss changes under small perturbations once simple rescaling effects are removed. The authors find that neural collapse can emerge during memorization, before generalization, while relative flatness changes sharply at the onset of generalization. Suppressing neural collapse does not stop the model from generalizing, so collapse is not necessary for grokking. Preventing relatively flat solutions does delay generalization, and in some settings it even creates delayed generalization that would not otherwise occur.

A different global perspective is proposed by \citet{zhang2025glass}. The paper treats the neural network as a physical system, with the parameters as degrees of freedom and the training loss as an energy-like variable, and studies the entropy of regions of parameter space indexed by training loss and test accuracy. Standard optimization quickly reaches a low training loss memorizing state and only later relaxes toward higher entropy states that generalize better. The authors argue that this transition does not involve an entropy barrier and therefore is incompatible with a strict first-order phase-transition interpretation.
A proposed optimizer supports this view by largely removing the delay before test generalization, while still reaching strong final generalization, and it does so without explicit weight norm constraints.

Finally, \citet{lee2024grokfast} turn the separation of timescales in grokking into a simple optimization method. The paper treats the gradient history of each parameter as a signal over training time and proposes that fast-varying components are linked to rapid overfitting, whereas slow-varying components are linked to delayed generalization. Their method amplifies the slow component by adding a low pass filtered version of recent gradients, implemented with a moving average or exponential moving average, to the current gradient. This modification greatly reduces the delay before generalization on modular multiplication and also improves training on image, language, and graph tasks. 

Despite progress, a unified explanation of grokking has not been proposed yet. Part of the difficulty may be that existing accounts often work at different levels of analysis, so they may describe different aspects of the same process rather than incompatible mechanisms. Clarifying how these description levels relate remains a central problem in the study of this phenomenon.

\subsection{Brain-Inspired Neural Models}

Brain-inspired neural models comprise several broad classes of approaches that draw on different aspects of neural computation. Some focus on neuron-level or synapse-level mechanisms, such as spikes, local plasticity, or structured connectivity, which change how signals propagate and how learning is implemented. Others reformulate learning through local inference and prediction, as in predictive coding frameworks. A third group introduces richer dynamical or biophysical structure, including dendritic nonlinearities, oscillatory or synchronization-based computation, and recurrent substrates such as reservoirs. Although these approaches differ in form, they share the aim of imposing stronger internal organization on the network, often to promote locality, sparsity, temporal dependencies, or more specialized representations.

The most developed branch is spiking neural networks (SNNs). A central challenge in recurrent spiking models is temporal credit assignment. Backpropagation through time solves this problem effectively, but it requires storing network states and propagating errors backward through the full sequence. \citet{bellec2020solution} address this limitation with e-prop, which expresses the update of each synapse as the product of two terms: a local eligibility trace and a learning signal. The eligibility trace summarizes how recent pre- and postsynaptic activity has influenced the hidden state, while the learning signal carries task-level error information. This produces an online learning rule that avoids full sequence replay and a global backward pass. 

\citet{fang2021deep} address an important limitation of spiking neural networks, i.e., the difficulty of training deep architectures reliably. Earlier SNNs often lost the main benefit of residual learning, because standard residual blocks do not transfer cleanly to spike-based computation and therefore still suffer from degradation and unstable gradients. The paper introduces spiking residual blocks, designed so that identity mappings remain easy to realize in a spiking setting. The residual branch is first converted into spikes, and the shortcut is then combined with those spikes through an element-wise operation in spike space. This change makes residual learning work for deep SNNs.

\citet{wang2023evolving} propose a different view of what should be learned in a recurrent spiking network. Instead of treating learning as the adjustment of synaptic weights, they treat it as the discovery of a sparse connectivity pattern. The model therefore learns, for each possible recurrent connection, a probability that the connection should exist. This makes the architecture itself part of the learned representation. The underlying claim is that much of the computational structure of a recurrent spiking network may lie in which neurons connect to which others, especially under the constraints of sparse neuromorphic computation. 

Predictive coding \citep{rao1999predictive} forms a second major family of brain-inspired models. \citet{golkar2022constrained} revisit one of its main biological difficulties: standard predictive-coding networks often rely on symmetric feedforward and feedback weights, along with tightly paired value and error units, assumptions that are difficult to reconcile with cortical circuitry. Their constrained formulation introduces covariance and decorrelation constraints that remove the need for exact symmetry between feedforward and feedback weights and also remove the need for one-to-one pairing of value and error units. This gives predictive coding a circuit interpretation that is closer to cortical organization, especially in models built from multicompartment neurons.

\citet{salvatori2024stable} reconsider how learning should be organized in predictive coding networks. In the standard formulation, the network first adjusts its latent activities through inference and only afterward updates the weights. Their incremental formulation removes this separation and allows parameter updates to occur throughout the inference process. Thus, inference and learning evolve together, which makes the method more stable.

\citet{ororbia2022lifelong} use predictive coding as a mechanism for continual learning. They combine local predictive coding updates with lateral competition and task selection dynamics, so the model can learn from a sequence of tasks without resetting its parameters. It is trained to reconstruct inputs and predict labels at the same time, which helps preserve useful internal structure as new tasks arrive. They show that online continual learning can emerge from local predictive dynamics rather than from replay or explicit protection of earlier parameters.

\citet{tang2023sequential} formulate predictive coding as a model of sequence memory instead of static inference. Their temporal predictive coding model learns the transition structure of a sequence, so that each state is trained to predict the next one. A stored sequence can then be recalled from a cue by running recurrent predictive updates through the learned transitions. The paper also shows that this dynamics is mathematically related to an asymmetric Hopfield network, with an additional decorrelating effect on stored patterns. The result also has a direct connection to theories of sequence memory in the hippocampus.

Other work draws inspiration from different mechanisms of neural computation. \citet{liu2024dendritic} start from the observation that biological dendrites do more than passively sum inputs before somatic activation; they can combine inputs nonlinearly within the neuron itself. The paper brings this idea into artificial networks by replacing standard linear integration with a quadratic integration rule. Each unit can represent pairwise interactions among features directly, rather than relying on later layers.

\citet{keller2023neural} replace recurrent updates with locally coupled oscillatory units whose hidden activity can organize into traveling waves. Computation is carried by the propagation and interaction of wave patterns across the recurrent substrate. This gives the network an explicit spatiotemporal structure, where local coupling imposes topography, and temporal information is represented through wave dynamics instead of a hidden state. Representation and memory emerge from organized collective activity.

The Kuramoto model \citep{kuramoto1975self} is an abstract model of rhythmic neural activity. Each neuron behaves like a simple oscillator with its own natural frequency, while coupling to other neurons pulls its phase toward that of its neighbors. In neuroscience terms, it captures how populations of oscillating neurons can move from independent rhythms and drifting phase relations to coherent collective activity through local interaction. 
Building on this idea, \citet{nguyen2024coupled} model a graph neural network (GNN) as a system of coupled dynamical units. In a standard GNN, each message-passing layer mixes the representation of a node with those of its neighbors. After many layers, this repeated mixing can wash out node-specific information, so representations of different nodes become too similar (over-smoothing). The paper interprets that collapse as the analogue of phase synchronization and then redesigns propagation to favor frequency synchronization instead. With the proposed message passing rule, nodes still evolve coherently across the graph, but they do not converge to the same representation.

Reservoir computing \citep{jaeger2001echo} uses a recurrent dynamical system with fixed internal connections to encode an input sequence into a high-dimensional state that evolves over time. The reservoir itself is not trained. Learning is confined to a readout that maps these state trajectories to the target output. The fixed recurrent dynamics are considered to generate a sufficiently expressive temporal representation, so the learning problem reduces to selecting an appropriate linear or weakly nonlinear readout.
\citet{gauthier2021next} show that the role of the reservoir does not actually require a random recurrent network. They replace it with a direct construction based on delayed copies of the input and nonlinear combinations of those delays. The source of the dynamics becomes explicit in the features themselves. This removes the random reservoir matrix, reduces the number of design choices that must be tuned, and makes the method easier to interpret mathematically. 

These brain-inspired models show that biological ideas can be used in artificial networks through several distinct routes. Some approaches modify signaling and temporal dynamics, others change how learning occurs locally, and others introduce interactions such as competition, synchronization, or richer intra-neuron computation. What connects these approaches is a common effort to impose more structure on the learning process and on the internal organization of the model. This is also the perspective adopted in the present work. We therefore investigate several biologically inspired mechanisms that have not been commonly used so far in artificial networks, including some associated not only with neurons, but also with other types of cells.

\section{BioNN Model Description}
\label{sec:model-description}

\subsection{Notation List}

Since the model contains a substantial number of parameters, we include the most relevant ones in Table \ref{table1}.

\begin{longtable}{>{\raggedright\arraybackslash}p{0.25\textwidth}
                  >{\raggedright\arraybackslash}p{0.30\textwidth}
                  >{\raggedright\arraybackslash}p{0.15\textwidth}
                  >{\raggedright\arraybackslash}p{0.30\textwidth}}
\caption{List of model parameters.}
\label{table1}\\

\toprule
Concept & Computation or value & Notation & Description \\
\midrule
\endfirsthead

\caption[]{List of model parameters (continued).}\\
\toprule
Concept & Computation or value & Notation & Description \\
\midrule
\endhead

\bottomrule
\endfoot

Input dimension & problem-dependent  & \(d\) & Number of input features. \\
Hidden dimensions &  problem-dependent & \(n\) & Number of hidden neurons. \\
Output dimension &  problem-dependent & \(q\) & Number of output neurons. \\
Layer input &  data-dependent & \(x\in\mathbb{R}^{d}\) & Input to the hidden layer. \\
Weight & learned & \(W\in\mathbb{R}^{n\times d}\) & Main hidden-layer weight matrix. \\
Bias & learned & \(b\in\mathbb{R}^{n}\) & Hidden-layer bias vector. \\
Preactivation & \(z=W_mx_g+b\) & \(z\in\mathbb{R}^{n}\) & Hidden-neuron value before ReLU. \\
Final hidden activity & \(a=\operatorname{ReLU}(\gamma\odot z-\Theta-\beta\ell)\) & \(a\in\mathbb{R}^{n}\) & Hidden activity after inhibition and final ReLU. \\
Readout weight & learned & \(W_y\in\mathbb{R}^{q\times n}\) & Output-layer weight matrix. \\
Readout bias & learned & \(b_y\in\mathbb{R}^{q}\) & Output-layer bias vector. \\
Model output & \(\hat{y}=W_ya+b_y\) & \(\hat{y}\in\mathbb{R}^{q}\) & Model prediction. \\
Modulator size & 32 & \(s\) & Dimension of the context vector. \\
Context vector & \(c=\tanh(Cx+b_c)\) & \(c\in\mathbb{R}^{s}\) & Learned context signal from the current input. \\
Modulator weight & learned & \(C\in\mathbb{R}^{s\times d}\) & Matrix that computes the context. \\
Modulator bias & learned & \(b_c\in\mathbb{R}^{s}\) & Bias vector for the context signal. \\
Input gate & \(g=\mathbf{1}+\alpha_i\tanh(Gc+b_g)\) & \(g\in\mathbb{R}^{d}\) & Multiplicative input-scaling vector. \\
Input gate weight & learned & \(G\in\mathbb{R}^{d\times s}\) & Matrix that computes the input gate. \\
Input gate bias & learned & \(b_g\in\mathbb{R}^{d}\) & Bias vector for the input gate. \\
Input gate strength & 0.6 & \(\alpha_i\) & Strength of context-based input scaling. \\
Gated input & \(x_g=x\odot g\) & \(x_g\in\mathbb{R}^{d}\) & Input after coordinate-wise gating. \\
Structural mask & state, initialized to ones & \(M\in\{0,1\}^{n\times d}\) & Binary mask over hidden-layer connections. \\
Masked weights & \(W_m=W\odot M\) & \(W_m\in\mathbb{R}^{n\times d}\) & Hidden weights after applying the mask. \\
Structural density & 0.35, 1 & \(\rho_m\) & Fraction of active entries in the mask. \\
Active connection count & \(k=\max(1,\lfloor \rho_mnd\rfloor)\) & \(k\) & Number of active entries in the mask. \\
Mask threshold & \(k\)-th largest value of \(|W_{ij}|\) & \(T_k\) & Weight-magnitude cutoff for mask update. \\
Gain vector & \(\gamma=\mathbf{1}+\alpha_g\tanh(Ac+b_a)\) & \(\gamma\in\mathbb{R}^{n}\) & Multiplicative response gain for hidden neurons. \\
Output gain weight & learned & \(A\in\mathbb{R}^{n\times s}\) & Matrix that computes hidden-neuron gain. \\
Output gain bias & learned & \(b_a\in\mathbb{R}^{n}\) & Bias vector for gain modulation. \\
Gain strength & 0.7 & \(\alpha_g\) & Strength of hidden-neuron gain modulation. \\
Homeostatic threshold & state, initialized to zero & \(\theta\in\mathbb{R}^{n}\) & Slow threshold adjusted by homeostasis. \\
Target activity & 0.12 & \(\rho^\star\) & Target activity rate for each hidden neuron. \\
Homeostasis rate & 0.01 & \(\eta\) & Step size for homeostatic threshold updates. \\
Threshold limit & 2 & \(\theta_{\max}\) & Maximum absolute homeostatic threshold. \\
Activity epsilon & \(10^{-4}, 10^{-7}\) & \(\varepsilon\) & Cutoff for counting a neuron as active. \\
Observed activity rate & \(\hat{\rho}_i=\frac{1}{B}\sum_b\mathbf{1}[a_i^{(b)}>\varepsilon]\) & \(\hat{\rho}\in\mathbb{R}^{n}\) & Batch activity frequency of hidden neurons. \\
Threshold shift & \(\Delta\theta=\alpha_t\tanh(Tc+b_t)\) & \(\Delta\theta\in\mathbb{R}^{n}\) & Fast context-dependent threshold shift. \\
Threshold shift weight & learned & \(T\in\mathbb{R}^{n\times s}\) & Matrix that computes fast threshold shifts. \\
Threshold shift bias & learned & \(b_t\in\mathbb{R}^{n}\) & Bias vector for fast threshold shifts. \\
Threshold strength &  0.6, 0.05  & \(\alpha_t\) & Strength of fast threshold modulation. \\
Total threshold & \(\Theta=\theta+\Delta\theta\) & \(\Theta\in\mathbb{R}^{n}\) & Threshold used in hidden-neuron activation. \\
Primary activity & \(p=\operatorname{ReLU}(\gamma\odot z-\Theta)\) & \(p\in\mathbb{R}^{n}\) & Hidden activity before lateral inhibition. \\
Lateral inhibition & \(\ell_i=(\sum_jp_j-p_i)/(n-1)\) & \(\ell\in\mathbb{R}^{n}\) & Inhibitory pressure from other hidden neurons. \\
Inhibition strength & 0.18 & \(\beta\) & Strength of lateral inhibition. \\
Minibatch size & problem-dependent & $B$ & Number of samples in a minibatch. \\
Batch activations & stack of final activities \(a\) over a minibatch & \(H\in\mathbb{R}^{B\times n}\) & Final hidden activations over a minibatch. \\
Centered activations & subtract column means from \(H\) & \(\bar{H}\) & Batch activations after column centering. \\
Normalized activations & normalize columns of \(\bar{H}\) & \(\hat{H}\) & Centered activations after column normalization. \\
Correlation matrix & \(R=\hat{H}^{\top}\hat{H}\) & \(R\in\mathbb{R}^{n\times n}\) & Pairwise hidden-neuron activity correlations. \\
Off-diagonal correlations & \(R_{\mathrm{off}}=R-\operatorname{diag}(\operatorname{diag}(R))\) & \(R_{\mathrm{off}}\) & Correlations excluding self-correlations. \\
Homeostasis regularization & \(10^{-4}\) & \(\lambda_h\) & Strength of the homeostatic threshold penalty. \\
Decorrelate strength & \(2 \cdot 10^{-3}\) & \(\lambda_d\) & Strength of the activation decorrelation loss. \\
Supervised loss & defined by the training objective & \({L}_y\) & Main prediction loss. \\
Homeostasis loss & \({L}_h=\lambda_h\frac{1}{n}\sum_i\theta_i^2\) & \({L}_h\) & Penalty on large homeostatic thresholds. \\
Decorrelation loss & \({L}_d=\lambda_d\operatorname{mean}(R_{\mathrm{off}}^2)\) & \({L}_d\) & Penalty on correlated hidden-neuron activity. \\
Total regularization loss & \({L}_r={L}_h+{L}_d\) & \({L}_r\) & Sum of BioNN regularization terms. \\
Total loss & \({L}={L}_y+{L}_r\) & \({L}\) & Prediction loss plus BioNN regularization. \\

\end{longtable}

\subsection{Overview of the Bio-Inspired Mechanisms}

A standard multilayer perceptron layer maps an input vector to a hidden representation by applying a linear combination of the input followed by a nonlinearity. In schematic form, this can be written as:

\begin{equation}
a = f(Wx + b)
\end{equation}
where \(x\) is the input, \(W\) and \(b\) are learned parameters, \(f\) is the activation function, and \(a\) is the hidden activity. This computation is fixed in the sense that the same mapping is applied to every input sample, apart from the values of the input itself.

The proposed model keeps the basic role of an MLP hidden layer, namely to transform the input into a hidden representation used by later layers. The difference is that this transformation is regulated by several biologically inspired mechanisms. These mechanisms do not replace the core computation. Instead, they modify how signals enter the layer, how strongly hidden neurons respond, which connections are active, how neurons compete, and how the activity of the layer is stabilized during training.

Before introducing these mechanisms, we note that the model includes a compact context representation derived from the input. For an input \(x \in \mathbb{R}^d\), the context vector is:

\begin{equation}
c = \tanh(Cx + b_c)
\end{equation}
where \(c \in \mathbb{R}^s\). This representation is used by some of the mechanisms as an additional source of information that modulates their behavior. Specifically, it controls input gating, gain modulation, and fast threshold modulation. Other mechanisms operate independently of it.

In a standard MLP, the input is passed directly to the linear map. In the proposed layer, the input may first pass through a regulation stage that can depend on the context representation. Some input coordinates may be amplified, while others may be suppressed. The resulting input is then processed through a sparse set of connections, whose active elements can change during training. After this step, the hidden neurons are further regulated by gain and threshold terms. Gain controls the strength of the response of each neuron, while thresholds control how much input is required for activation. The layer then applies competition among hidden neurons through lateral inhibition, which operates on the candidate activity after the nonlinearity and suppresses diffuse co-activation.

\begin{figure}[ht]
\centering
\begin{tikzpicture}[
    mechanism/.style={
        draw,
        rectangle,
        rounded corners,
        very thick,
        align=center,
        minimum height=8mm,
        minimum width=34mm
    },
    state/.style={
        draw,
        rectangle,
        dashed,
        align=center,
        minimum height=8mm,
        minimum width=30mm
    },
    arrow/.style={
        -{Latex[length=2.2mm]},
        thick
    },
    dashedarrow/.style={
        -{Latex[length=2.2mm]},
        thick,
        dashed
    }
]

\node[state] (context) at (0,4.8)
    {Context\\from input};

\node[mechanism] (inputgating) at (-5,2.8)
    {Input\\gating};

\node[mechanism] (gain) at (0,2.8)
    {Gain\\modulation};

\node[mechanism] (threshold) at (5,2.8)
    {Threshold\\modulation};

\node[state] (gatedinput) at (-5,0.6)
    {Gated\\input};

\node[mechanism] (structural) at (-5,-1.4)
    {Structural\\plasticity};


\node[state, text opacity=0] (preactivation) at (-1.2,-1.4)
    {A\\B};
\node at (preactivation.center)
    {Pre-activation};

\node[state] (candidate) at (2.8,-1.4)
    {Candidate\\activity};

\node[mechanism] (inhibition) at (2.8,-3.4)
    {Lateral\\inhibition};

\node[state] (finalactivity) at (2.8,-5.4)
    {Final hidden\\activity};


\node[mechanism, text opacity=0] (homeostasis) at (6.6,-3.4)
    {A\\B};
\node at (homeostasis.center)
    {Homeostasis};

\node[mechanism] (decorrelation) at (-1.2,-5.4)
    {Activation\\Decorrelation};

\draw[arrow] (context) -- (inputgating);
\draw[arrow] (context) -- (gain);
\draw[arrow] (context) -- (threshold);

\draw[arrow] (inputgating) -- (gatedinput);
\draw[arrow] (gatedinput) -- (structural);
\draw[arrow] (structural) -- (preactivation);
\draw[arrow] (preactivation) -- (candidate);

\draw[arrow]
    (gain.south) -- ++(0,-1.0) -| (candidate.north west);

\draw[arrow]
    (threshold.south) -- ++(0,-1.0) -| (candidate.north east);

\draw[arrow]
    (homeostasis.north) |- (candidate.east);

\draw[arrow] (candidate) -- (inhibition);
\draw[arrow] (inhibition) -- (finalactivity);

\draw[dashedarrow]
    (finalactivity.east) -| (homeostasis.south);

\draw[dashedarrow]
    (decorrelation.east) -- (finalactivity.west);

\end{tikzpicture}

\caption{Overview of a BioNN hidden layer. The input-derived context controls input
gating, gain modulation, and fast threshold modulation. Structural plasticity determines
the connectivity mask used in the affine transformation. Gain modulation and the fast
and homeostatic thresholds regulate primary activity, which provides the input to lateral
inhibition before the final hidden activity is computed. During training, the final activity
is used to update the homeostatic threshold and to compute the activation-decorrelation
loss. Solid arrows indicate forward or control dependencies, while dashed arrows indicate
training-time adaptation or regularization effects.}
\label{fig:bionn-overview}

\end{figure}
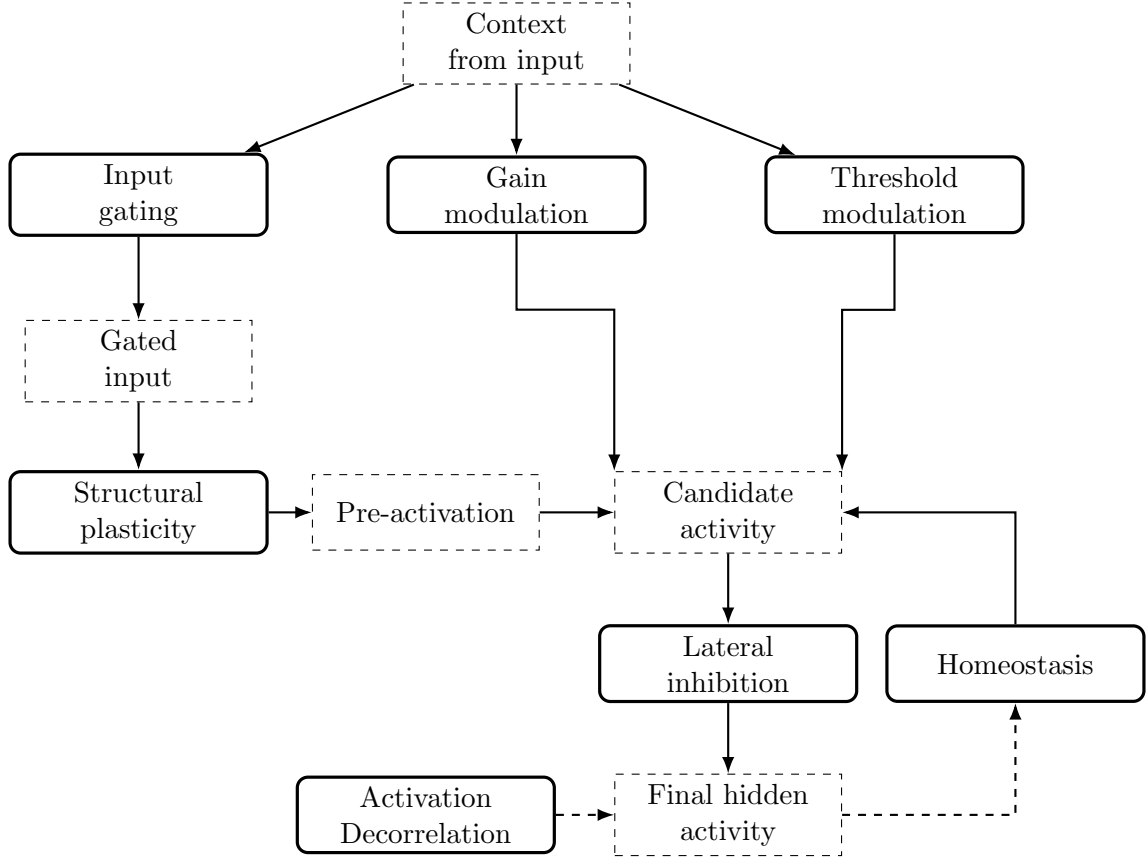

Figure~\ref{fig:bionn-overview} summarizes the organization of the regulated hidden layer and the points at which the individual mechanisms act.

This diagram separates two roles of the model. The first role is fast, sample-dependent regulation. The context representation influences only the mechanisms that use it, such as input gating, gain modulation, and fast threshold modulation. The second role is slower training-time regulation. During learning, Homeostasis adjusts the long-term activation thresholds of hidden neurons, Structural Plasticity changes which connections participate in the effective circuit, and a Decorrelation loss term discourages hidden neurons from repeatedly encoding the same information. Lateral inhibition acts within the forward computation and introduces competition among the hidden neurons based on their current activity.

A compact form of the computation is:

\begin{equation}
x_g = x \odot g
\end{equation}
\begin{equation}
z = W_m x_g + b
\end{equation}
\begin{equation}
\Theta = \theta + \Delta\theta
\end{equation}
\begin{equation}
p = \operatorname{ReLU}(\gamma \odot z - \Theta)
\end{equation}
\begin{equation}
a = \operatorname{ReLU}(\gamma \odot z - \Theta - \beta \ell)
\end{equation}
where \(g\), \(\gamma\), and \(\Delta\theta\) can depend on the context vector \(c\), \(W_m = W \odot M\) is the structurally masked weight matrix, \(\theta\) is the slowly updated homeostatic threshold state, \(p\) is the primary activity before inhibition, and \(\ell\) is the lateral inhibitory pressure. The dependence on \(c\) applies only to the mechanisms that use the context representation.

The final hidden activity is passed to the readout layer:

\begin{equation}
\hat{y} = W_y a + b_y.
\end{equation}

The following subsections define the individual mechanisms and their updates.

\subsection{Input Gating}

In biological neurons, inputs arriving through different dendritic branches do not contribute equally to the final response. Local circuit mechanisms modulate the influence of specific input pathways. Inhibitory interneurons can suppress activity on particular dendritic branches, while disinhibitory circuits can selectively restore the influence of those pathways. This form of pathway-specific modulation allows the neuron to adjust how strongly different inputs affect its output, depending on context \citep{YangMurrayWang2016}.

This observation motivates the input gate in the proposed model. The gate applies a context-dependent reweighting of input features before the main hidden-layer computation. It increases the contribution of some features and reduces the contribution of others. This reflects the idea that the response of a neuron depends not only on synaptic weights but also on mechanisms that regulate how inputs are integrated under different conditions. In computational terms, the gate modifies the effective input representation. As a result, the same hidden neuron can produce different responses because it receives a version of the input that has been adjusted according to the current context.

This design is consistent with the view of dendrites as active computational structures. Dendrites support structured processing of incoming signals, including nonlinear integration and pathway-specific modulation. When inputs from different sources arrive on separate dendritic branches, local inhibitory and disinhibitory mechanisms can selectively adjust their influence \citep{StuartSpruston2015}. The input gate provides a compact abstraction of this form of selective integration within a standard neural network layer.

Formally, the input gate rescales each input coordinate before the main affine transformation:

\begin{equation}
g = 1 + \alpha_i \tanh(Gc + b_g)
\end{equation}
\begin{equation}
x_g = x \odot g.
\end{equation}

Here, \(\alpha_i\) controls the strength of input gating, and \(g \in \mathbb{R}^d\) contains one scaling factor per input coordinate. Values of \(g\) greater than one amplify the corresponding input dimensions, while values smaller than one attenuate them. The layer therefore transforms the original input \(x\) into a reweighted input \(x_g\) that depends on the context.

The trainable parameters of the input gate are \(G\) and \(b_g\), while \(C\) and \(b_c\) define the shared context representation introduced in the previous subsection. All parameters are learned through backpropagation. The input-gate head is initialized to zero, which yields an initial gate equal to the neutral vector \(\mathbf{1}\). At initialization, the gate leaves the input unchanged. During training, it learns to introduce context-dependent reweighting.

\subsection{Structural Plasticity}

In the brain, learning can modify both the strength of existing synapses and the underlying physical connectivity. Neural circuits can form new dendritic spines (small protrusions that serve as postsynaptic sites), eliminate existing ones, add or remove axonal boutons (presynaptic swellings where neurotransmitter release occurs), and reorganize local branches over time. These processes determine which neurons are effectively connected and which pathways support signal flow. Structural plasticity refers to this continuous reconfiguration of connectivity. It is observed across multiple timescales and functional contexts: during development, when large-scale circuits are formed and refined; during experience-dependent learning, when activity patterns reshape connectivity; and in the adult cortex, where experience continues to remodel connections in support of adaptation and memory \citep{FuZuo2011,BernardinelliMullerNikonenko2014,BoschHayashi2014}.

This biological mechanism motivates a view of learning in which improvement arises from changes in the connectivity of the model. Biological networks are sparse, selective, and dynamic. A neuron does not maintain a fixed set of effective neighbors over time. Connections that consistently contribute to useful computation can stabilize, weak connections can disappear, and new candidate connections can emerge when repeatedly recruited by experience. This perspective supports a computational mechanism that can prune, preserve, and reorganize connectivity during training \citep{ChklovskiiMelSvoboda2004,ButzVanOoyen2014}.

Early in learning, many pathways may be available. With experience, the system retains the pathways that repeatedly support useful computation and reduces the pathways that are rarely used or that interfere with stable processing. Studies of cortical spine dynamics describe this pattern: some synaptic structures are transient, some become stabilized through experience, and the resulting circuit becomes more specialized over time.

The proposed model captures this principle through a structural mask that determines which connections participate in the effective computation. Formally, structural plasticity introduces a binary mask into the weight matrix:

\begin{equation}
W_m = W \odot M
\end{equation}
\begin{equation}
z = W_m x_g + b.
\end{equation}

Here \(M \in \{0,1\}^{n \times d}\) is the structural mask. If \(M_{ij}=1\), the connection from input coordinate \(j\) to hidden neuron \(i\) is active. If \(M_{ij}=0\), that connection is absent from the effective circuit. The model still stores the dense trainable matrix \(W\), but the forward pass uses only the entries selected by \(M\).

This mechanism makes the effective layer sparse. The active connection density is controlled by \(\rho_m\). The mask is initialized with the specified density \(\rho_m\). During training, the mask can be refreshed by a magnitude-based rule. The rule keeps the top-\(k\) entries of \(\lvert W_{ij} \rvert\), where
$k = \max\left(1,\left\lfloor \rho_m nd \right\rfloor\right)$.

Let \(T_k\) denote the magnitude threshold associated with the top-\(k\) entries. The structural update can be written as:

\begin{equation}
M_{ij} \leftarrow \mathbf{1}\left[\lvert W_{ij}\rvert \geq T_k\right].
\end{equation}

Large weights remain in the effective circuit, while weak weights disappear from the forward computation.

The structural density also determines the strength of the structural constraint. When
\(\rho_m<1\), only a fraction of the connections participates in the effective forward
computation, and the periodic mask update can change which connections form the active
circuit. When \(\rho_m=1\), all connections remain active, so the effective layer is dense
and the structural mask does not impose connection selection. The latter setting therefore
provides a useful limiting case for separating the effect of explicit structural sparsification
from the presence of the structural plasticity mechanism itself.

The training distinction is important. The matrix \(W\) is learned by gradient descent. The mask \(M\) is updated by an explicit rule outside gradient descent. The structural update is applied periodically during training, with the update schedule specified by the experiment.

In a sparse setting, different patterns of connectivity correspond to different effective circuits, and selecting among them changes the space of computations that the network can realize. This perspective treats connectivity as a resource that can be allocated and reallocated during training. The structural mask provides a simple abstraction of this idea by allowing the model to modify which connections participate in the forward computation, while the dense weight matrix is retained as the underlying parameter matrix. In this way, learning operates over both parameter values and circuit structure, so changes in connectivity can reshape network function.

\subsection{Gain Modulation}

In the brain, the strength of a neuron response can change even when the preferred input pattern of the neuron remains the same. For example, a visual neuron may continue to prefer the same orientation or spatial location, while its response becomes stronger or weaker depending on attention, arousal, motor state, or neuromodulatory input. This effect is known as gain modulation. It changes the amplitude or sensitivity of the response while preserving the basic selectivity of the neuron. Gain modulation has therefore been described as a major computational principle in the central nervous system \citep{SalinasThier2000,SalinasThier2001}.

Several biological mechanisms can change gain. Cortical neurons are affected by background excitation and inhibition, local circuit state, and neuromodulators such as acetylcholine and noradrenaline. These factors change how strongly the firing rate of a neuron increases as input drive increases. The same input can therefore produce a weak response in one state and a stronger response in another. Reviews of cortical gain \citep{FergusonCardin2020} describe this form of regulation in attention, arousal, motor activity, and neuromodulatory control, where different systems can adjust cortical processing through related circuit mechanisms. \citet{MunnEtAl2021} provide a systems-level counterpart to this idea. They treat arousal as a low-dimensional modulatory signal that changes cortical neural gain and thereby changes the dynamical regime of a much larger network.

Thus, gain changes the strength of the response of the neuron within the current context. This is an abstraction of state-dependent excitability, meaning that neural circuits can keep the same basic organization while processing signals with different levels of amplification.

Formally, gain modulation changes how strongly each hidden neuron responds to its preactivation:

\begin{equation}
\gamma = 1 + \alpha_g \tanh(Ac + b_a).
\end{equation}

Here \(\alpha_g\) controls the strength of gain modulation, and \(\gamma \in \mathbb{R}^n\) has one entry per hidden neuron. The vector \(\gamma\) changes how excitable each neuron is for the current instance.

A large gain makes a neuron respond more strongly to the same input. A small gain makes it less responsive. The parameters \(A\) and \(b_a\) are learned by backpropagation. The gain-modulation head is initialized to zero, so the initial gain is the neutral vector \(\mathbf{1}\).

With gain modulation, the network can keep the same representational components while changing how strongly they contribute under different contexts.

\subsection{Threshold Modulation}

In the brain, the firing threshold of a neuron is variable. It changes with the recent state of the cell, the rate of incoming depolarization, background inhibitory and excitatory activity, and slower adjustments in intrinsic excitability. Depending on the context, a neuron can become easier or harder to activate. The same input pattern can therefore produce a different output when the internal activation threshold of the cell has changed \citep{AzouzGray2000,WesterContreras2013}.

The activation threshold of a cortical neuron depends on the recent input context. Rapidly accumulating input can lower the effective firing criterion, so coordinated evidence is more likely to activate the neuron \citep{AzouzGray2000}. Another study extends this idea from coincidence detection to sensitivity regulation. Adaptive threshold changes reduce the effect of slow input changes and preserve sensitivity to rapid input fluctuations, so the neuron adjusts how strongly it responds to different input dynamics \citep{AzouzGray2003}. There is also a slower biological process behind threshold regulation. The axon initial segment, where spikes are typically initiated, can change its structure and function over time. This plasticity affects neuronal excitability and is closely linked to homeostatic regulation, presented in the next subsection. The brain can therefore retune how easily a neuron fires when network activity has been too strong or too weak for a sustained period. Threshold modulation therefore has two biological justifications: fast state-dependent changes in activation threshold and slower adaptive adjustments that stabilize network function \citep{Kuba2010,YamadaKuba2016,PetersenEtAl2017}.

The proposed model separates these two timescales. Fast threshold modulation adds a sample-dependent threshold shift:

\begin{equation}
\Delta\theta = \alpha_t \tanh(Tc + b_t).
\end{equation}

The total threshold used by the hidden neuron is:

\begin{equation}
\Theta = \theta + \Delta\theta.
\end{equation}

Here, \(\alpha_t\) controls the strength of fast threshold modulation. The component \(\Delta\theta\) changes from sample to sample through the context vector \(c\), whereas the homeostatic component \(\theta\) changes more slowly through the activity-dependent update described in the next subsection.

The parameters \(T\) and \(b_t\) are learned by backpropagation. The threshold-modulation head is initialized to zero, so the initial fast threshold shift is \(\Delta\theta=\mathbf{0}\). The mechanism gives the current input a direct influence on how easy it is to activate each hidden neuron.

\subsection{Homeostasis}

Homeostasis is a biological mechanism for keeping neural activity within a stable and useful range. Neural circuits continue to change during learning, and these changes can push some neurons toward excessive activity or near silence. Homeostatic plasticity counteracts this drift by lowering responsiveness when activity is high and increasing responsiveness when activity is low. This gives the circuit a slow stabilizing process that preserves its ability to learn while synaptic and structural changes continue to take place \citep{Turrigiano2012,Davis2013}.

Homeostatic synaptic plasticity provides a synaptic mechanism for this stabilizing principle. A central example is synaptic scaling, where a neuron adjusts the overall strength of its excitatory synapses in response to sustained changes in firing activity. This can raise or lower the average activation level of the neuron while preserving much of the relative structure of the learned synaptic pattern \citep{Turrigiano2008}. More generally, homeostatic synaptic plasticity covers related feedback mechanisms that adjust synaptic strengths when excitation or inhibition remains away from a stable range \citep{PozoGoda2010}. The computational idea is that learning can shape what a neuron responds to, and a slower regulatory process can control how often the neuron responds.

The proposed model applies this idea through a homeostatic threshold for each hidden neuron. During training, the model estimates how often each hidden neuron activates across instances. If a neuron activates too often, its threshold slowly increases. If it activates too rarely, its threshold slowly decreases. The homeostatic update acts directly on the threshold state rather than on the weight vector that defines neuronal selectivity. The result is a slow compensatory process that keeps the hidden layer in a useful activity range while gradient-based learning continues to organize the representation. This matches the computational view that associative learning needs stabilizing mechanisms across multiple timescales \citep{ZenkeGerstner2017}.

The proposed model gives each hidden neuron a slow threshold state
$\theta \in \mathbb{R}^{n}$.
The vector \(\theta\) is a state variable rather than a parameter learned by gradient descent. For a minibatch of size \(B\), the model measures how often neuron \(i\) is active:

\begin{equation}
\hat{\rho}_i =
\frac{1}{B}
\sum_{r=1}^{B}
\mathbf{1}\!\left[a_i^{(r)} > \varepsilon\right].
\end{equation}

Here, \(\varepsilon\) is the activity cutoff and \(r\) indexes examples in the minibatch. The homeostatic threshold is then updated according to:

\begin{equation}
\theta_i \leftarrow
\operatorname{clip}
\left(
\theta_i + \eta(\hat{\rho}_i - \rho^\star),
-\theta_{\max},
\theta_{\max}
\right).
\end{equation}

Here, \(\eta\) is the homeostatic update rate, \(\rho^\star\) is the target activity rate, and \(\theta_{\max}\) bounds the magnitude of the threshold state.

The meaning of this update is that if a neuron fires too often, its threshold increases. If it fires too rarely, its threshold decreases. This counteracts the tendency of the population to collapse into a few persistently active neurons and many persistently silent neurons. The update is performed explicitly from the final hidden activity \(a\) and does not use gradient descent.

\subsection{Lateral Inhibition}

Lateral inhibition provides a biological mechanism for selection through competition. When activity appears in one part of a neural population, inhibitory interactions can suppress neighboring or competing activity, so the strongest response becomes more distinct and weaker alternatives are reduced. This principle is especially clear in sensory systems, where inhibition can sharpen the representation of local differences rather than allowing broad, diffuse activation. In cortical circuits, lateral inhibition can also support winner-take-all behavior and act as a spatial and temporal high-pass filter for sensory input \citep{FanEtAl2020}.

The classical example comes from early visual processing in the retina. Horizontal cells pool signals from surrounding photoreceptors and send inhibitory feedback to photoreceptors near the center of the receptive field. This creates center-surround organization, which makes local contrast easier to represent and contributes to edge enhancement, color discrimination, and light adaptation \citep{KramerDavenport2015}. In the cortex, the role of inhibition is more complex: lateral inhibition has often been proposed as a mechanism for sharpening stimulus selectivity, although some selectivity effects can also arise from feedforward structure, spike threshold, and response nonlinearities \citep{PriebeFerster2008}.

This biological mechanism motivates a competitive hidden layer. If several neurons respond to similar evidence, the circuit does not simply let all of them remain equally active. Local inhibition pushes the representation toward a cleaner allocation, where the most strongly supported response becomes dominant and nearby alternatives are reduced. This process makes the code more selective and less diffuse. It also helps prevent broad, redundant co-activation across similar neurons.

In the proposed model, lateral inhibition is applied after gain and thresholds have been applied. The layer first computes a candidate activity:

\begin{equation}
p = \operatorname{ReLU}(\gamma \odot z - \Theta).
\end{equation}

It then computes inhibitory pressure from the candidate activity. In the implementation used here, inhibition is global across the hidden layer: the inhibitory pressure on neuron \(i\) is the mean primary activity of all other hidden neurons:

\begin{equation}
\ell_i =
\frac{\sum_{j=1}^{n} p_j - p_i}{n - 1}.
\end{equation}

The final hidden activation is:

\begin{equation}
a = \operatorname{ReLU}(\gamma \odot z - \Theta - \beta \ell),
\end{equation}
where \(\beta\) controls the strength of lateral inhibition.

This mechanism creates competition among neurons active at the same time. Each neuron competes against the current activity of the rest of the population. Diffuse activation becomes harder to maintain, while more selective patterns become easier to stabilize. Lateral inhibition introduces no separate learned weight matrix. The scalar \(\beta\) is a model parameter fixed by the experimental configuration, while gradients still propagate through the inhibition computation to the trainable parameters of the layer.

\subsection{Activation Decorrelation}

Neighboring neurons often receive overlapping input, and their responses can therefore become redundant. Efficient-coding theory treats this redundancy as a computational problem: a population code should use its neurons to carry complementary information, rather than many copies of the same signal. In vision, this principle is especially relevant because natural scenes contain strong regularities in space and time, and early sensory circuits must encode these regularities under limited bandwidth and metabolic constraints \citep{MaheswaranathanEtAl2023,ManookinSimoncelli2023}.

The retina provides direct evidence for this principle. Retinal ganglion cell spike trains are less correlated than the corresponding locations in naturalistic visual input, which supports the view that retinal processing reduces redundancy before visual information reaches the brain \citep{PitkowMeister2012}. Retinal responses also adapt to the recent statistical structure of the stimulus: predictable patterns are attenuated, while deviations from those patterns remain more salient. This gives retinal coding a dynamic predictive component, where the output emphasizes information that is less expected from recent input history \citep{HosoyaBaccusMeister2005}.

Two mechanisms are especially relevant for the proposed model. First, adaptation can decorrelate activity over time. In visual cortical neurons, intrinsic adaptive dynamics can reduce temporal correlations in the output spike train when the input contains slow, naturalistic temporal correlations \citep{WangEtAl2003}. Second, inhibition can decorrelate activity across a population. Experiments in neocortical circuits show that inhibition can reduce shared fluctuations among nearby neurons \citep{SippyYuste2013}, and spiking models of V1 show that inhibitory interneurons can suppress predictable excitatory spikes, leading to sparse and decorrelated representations \citep{KingEtAl2019}. 

The proposed model abstracts this principle as a decorrelation loss on hidden activations: hidden neurons are encouraged to specialize rather than repeatedly encode the same variation across instances.

The proposed model implements activation decorrelation through the loss rather than through the forward map itself. Let
$H \in \mathbb{R}^{B \times n}$
be the matrix of final hidden activations over a minibatch of size \(B\). The activity of each hidden neuron is centered across the minibatch to obtain \(\bar{H}\), and each column is then normalized to obtain \(\hat{H}\). The hidden-neuron correlation matrix is:

\begin{equation}
R = \hat{H}^{\top}\hat{H}.
\end{equation}

The diagonal entries represent self-correlations and are removed:

\begin{equation}
R_{\mathrm{off}} =
R - \operatorname{diag}(\operatorname{diag}(R)).
\end{equation}

The decorrelation loss is:

\begin{equation}
L_d =
\lambda_d\,\operatorname{mean}\!\left(R_{\mathrm{off}}^2\right),
\end{equation}
where \(\lambda_d\) controls the strength of activation decorrelation.

If two hidden neurons repeatedly fire in very similar ways across the batch, their correlation contributes to the penalty. Training then pushes the model toward less redundant hidden codes. This mechanism is learned in the usual way for a regularizer. The penalty is added to the task loss, and gradients from the combined loss update the trainable parameters.

\section{Experimental Setup}

\subsection{The Grokking Problems Under Study}

The experimental study uses two binary classification problems that are well suited for analyzing grokking and the internal structure of neural networks. Both problems contain a simple underlying rule, but the rule is difficult to identify from finite data because it is embedded in a much larger space of irrelevant variation. This makes them useful test cases for separating memorization from genuine generalization. A model can fit the training labels by using accidental correlations in the sample, but good test performance requires it to recover the compact structure that generated the labels. The two tasks are sparse parity, originally studied as a grokking problem by \citep{merrill2023tale}, and noisy XOR classification, originally studied in the context of benign overfitting and grokking by \citep{xu2023benign}.

\subsubsection{Sparse Parity}

The sparse parity task is a binary classification problem on sign vectors. Each input is a vector
$\mathbf{x} = (x_1,\ldots,x_{40}) \in \{-1,+1\}^{40}$.
The target label is determined by the product of a small subset of input coordinates. In the experiments, the relevant subset consists of the first three coordinates, so the label is
$y = x_1 \cdot x_2 \cdot x_3 \in \{-1,+1\}$.
The remaining coordinates are random signs that do not affect the label. They make the input high-dimensional, but they contain no information about the rule.

This problem is important for grokking because it separates the size of the input space from the size of the true computation. The rule itself is compact: only three coordinates are relevant. However, the model receives forty coordinates, and most of them act as distractors. A network trained on a finite sample can initially fit the training data through a dense, sample-specific solution that uses many incidental patterns in the random coordinates. Such a solution can achieve high training accuracy while it fails to capture the true parity rule. Generalization requires the model to identify the small set of relevant coordinates and to implement the multiplicative sign interaction among them.

The main difficulty is that parity is not an additive or linearly separable rule in the raw input coordinates. Each relevant coordinate alone is uninformative about the label: flipping any one of the three relevant signs flips the target, but no single coordinate determines the class. The useful signal is contained in a higher-order interaction among the relevant coordinates. This makes sparse parity a natural setting for studying whether a neural network forms an internal circuit that represents the true rule, rather than merely fitting the sample.

For this reason, sparse parity is also useful for structural analysis. If a network groks the task, one expects its hidden representation to become more organized around the true rule. In particular, the learned computation should depend on a relatively small subset of useful neurons or features, rather than on a broad collection of weak memorizing components. Structural indicators such as activation sparsity, effective subnetwork size, neuron specialization, and support overlap are therefore meaningful in this setting. They can reveal whether the transition to generalization coincides with the emergence of a compact internal representation.

\subsubsection{Noisy XOR Clusters}

The noisy XOR cluster task is a binary classification problem in a high-dimensional Euclidean space. Each input belongs to one of four latent clusters. Two clusters correspond to the positive class, and two clusters correspond to the negative class. In the experiments, each input is \(x \in \mathbb{R}^{40000}\).

The informative structure is concentrated in two signal directions. A clean label \(\tilde{y} \in \{-1,+1\}\) first selects which signal direction is relevant. An independent sign then selects one of the two cluster centers on that direction. Thus, positive examples are centered around the two opposite locations on one axis, while negative examples are centered around the two opposite locations on another axis. The remaining coordinates contain isotropic noise. The two cluster-center vectors have norm \(2.5\sqrt{d/N_{\mathrm{train}}}\). Standard Gaussian noise is added in all \(d=40{,}000\) dimensions, and the resulting input vector is divided by \(\sqrt d\). Training labels are independently flipped with probability 0.05, while test labels remain clean.

The training labels include a small amount of random label noise. In the experiments, a fraction of the training labels is flipped, while the test labels remain clean. This creates a setting in which the model must learn the latent cluster structure despite corrupted supervision. The true classification rule is tied to the axis that generated the example, not to the accidental identity of individual training points.

This task is important for grokking because it creates a gap between fitting and understanding. A sufficiently expressive model can fit the noisy training labels, including the flipped ones. Such fitting does not by itself imply that the model has recovered the cluster-level rule. Good generalization requires the model to capture the latent XOR structure rather than rely on sample-specific correlations; a useful representation is therefore expected to reflect the four latent groups. The problem therefore supports a temporal separation between memorization and generalization: the model may first fit the sample and only later organize its hidden representation around the true cluster geometry.

The main difficulty comes from the combination of nonlinearity, high dimensionality, and label noise. The four clusters are arranged so that the class is not represented by a single linear direction through the whole input space. A classifier must distinguish the two positive clusters from the two negative clusters, even though each class is split across opposite centers. The large number of noisy coordinates also creates many directions in which a model can fit finite-sample fluctuations. Label noise adds another source of difficulty because some training labels explicitly contradict the latent rule.

The noisy XOR cluster problem is therefore well suited for studying internal structural indicators. The hidden representation can be evaluated according to whether it separates the four latent clusters, whether neurons specialize to particular groups, whether same-group samples activate similar supports, and whether different groups use distinct hidden-neuron patterns. These indicators are directly connected to the central question of grokking in this setting: whether late generalization corresponds to a qualitative reorganization of the hidden layer from sample-specific fitting toward a representation of the latent data-generating structure.

\subsection{Network Structure Indicators}

All indicators are computed from the hidden activations of the trained network. For \(N\) evaluated samples and \(n\) hidden neurons, let \(H \in \mathbb{R}^{N \times n}\) denote the hidden activation matrix. A hidden neuron \(u\) is treated as active for sample \(i\) when \(|H_{iu}| > \varepsilon\). The activity cutoff is \(\varepsilon = 10^{-4}\) for sparse parity and \(\varepsilon = 10^{-7}\) for noisy XOR. The Jaccard metrics use 8192 sampled pairs, and a numerical cutoff \(\delta = 10^{-12}\) is used where required. For sparse parity, the groups are the two target classes. For noisy XOR, the groups are the four latent clusters \(+\mu_1,-\mu_1,+\mu_2,-\mu_2\).

\paragraph{Lifetime sparsity.}
Lifetime sparsity is the average fraction of samples for which one hidden neuron is inactive. High values mean that neurons are used only on limited parts of the data, whereas low values indicate broad activity across samples. For each neuron or unit \(u\):

\begin{equation}
\ell_u =
\frac{1}{N}
\sum_{i=1}^{N}
\mathbf{1}\!\left[|H_{iu}| \leq \varepsilon\right].
\end{equation}

Lifetime sparsity is:

\begin{equation}
S_{\mathrm{life}} =
\frac{1}{n}
\sum_{u=1}^{n}
\ell_u.
\end{equation}

\paragraph{Sample Hoyer sparsity.}
Sample Hoyer sparsity is a magnitude-sensitive sparsity score for each sample activation vector. For \(h_i \in \mathbb{R}^{n}\), the \(i\)-th row of \(H\):

\begin{equation}
\operatorname{Hoyer}(h_i)
=
\frac{
\sqrt{n} - \|h_i\|_1/\|h_i\|_2
}{
\sqrt{n} - 1
}.
\end{equation}

The reported score is:

\begin{equation}
S_{\mathrm{Hoyer,sample}}
=
\frac{1}{N}
\sum_{i=1}^{N}
\operatorname{Hoyer}(h_i).
\end{equation}

A representation dominated by one or a few neurons has a high score, while activity distributed more evenly across neurons has a lower score. If \(\|h_i\|_2 \leq \delta\), the score is defined as 1.

\paragraph{Neuron Hoyer sparsity.}
Neuron Hoyer sparsity applies the same measure to the activation profile of each hidden neuron across all samples. Let \(h^{(u)} \in \mathbb{R}^{N}\) denote column \(u\) of \(H\). The metric is:

\begin{equation}
S_{\mathrm{Hoyer,neuron}}
=
\frac{1}{n}
\sum_{u=1}^{n}
\frac{
\sqrt{N} - \|h^{(u)}\|_1/\|h^{(u)}\|_2
}{
\sqrt{N} - 1
}.
\end{equation}

High values indicate that a typical neuron is strongly active on only a restricted subset of samples.

\paragraph{Dead-neuron fraction.}
Dead-neuron fraction measures the proportion of hidden neurons that are effectively unused. Let \(\mathcal{I}_g\) be the set of samples belonging to group \(g\), for \(g=1,\ldots,K\), where \(K\) is the number of groups. For neuron \(u\), define its mean absolute activity in group \(g\) as:

\begin{equation}
\mu_{ug}
=
\frac{1}{|\mathcal{I}_g|}
\sum_{i \in \mathcal{I}_g}
|H_{iu}|.
\end{equation}

Its total activity across groups is:

\begin{equation}
m_u =
\sum_{g=1}^{K}
\mu_{ug}.
\end{equation}

A neuron is considered dead when \(m_u \leq \delta\). The dead-neuron fraction is:

\begin{equation}
S_{\mathrm{dead}}
=
\frac{
\#\{u : m_u \leq \delta\}
}{
n
}.
\end{equation}

A value near zero means that almost every hidden neuron contributes somewhere in the data.

\paragraph{Mean absolute correlation.}
Mean absolute correlation measures redundancy among hidden neurons. Each neuron's activation profile is centered across samples:

\begin{equation}
\widetilde{H}_{iu}
=
H_{iu}
-
\frac{1}{N}
\sum_{j=1}^{N}
H_{ju}.
\end{equation}

Neurons with negligible variance are excluded. If \(K_c\) nonconstant neurons remain, the metric is:

\begin{equation}
C_{\mathrm{abs}}
=
\frac{1}{K_c(K_c-1)}
\sum_{u \neq v}
\left|
\operatorname{Corr}
\!\left(
\widetilde{H}_{\cdot u},
\widetilde{H}_{\cdot v}
\right)
\right|.
\end{equation}

High values indicate greater redundancy. Because absolute correlation is used, strong negative correlations also contribute to the measure.

\paragraph{Effective rank.}
Effective rank measures the dimensionality of the hidden representation. Let \(s_1,\ldots,s_r\) be the nonzero singular values of the centered activation matrix and define:

\begin{equation}
p_k =
\frac{s_k}{\sum_j s_j}.
\end{equation}
The singular-value entropy is
\begin{equation}
S_{\mathrm{sv}}
=
-\sum_k
p_k \log(p_k+\delta).
\end{equation}

The normalized effective rank is:

\begin{equation}
R_{\mathrm{eff}}
=
\frac{
\exp(S_{\mathrm{sv}})
}{
\min(N,n)
}.
\end{equation}

The symbol \(S_{\mathrm{sv}}\) is used for the singular-value entropy to avoid reusing \(H\), which already denotes the hidden activation matrix. High values indicate that the representation occupies many independent directions in hidden space; low values indicate concentration in a smaller effective subspace.

\paragraph{Specialization.}
Specialization measures how group-specific the hidden neurons are. Using the group-conditioned activities \(\mu_{ug}\), let us define:

\begin{equation}
p_{ug}
=
\frac{
\mu_{ug}
}{
\sum_{h=1}^{K}
\mu_{uh}
}.
\end{equation}

The specialization of neuron \(u\) is:

\begin{equation}
\operatorname{Spec}_u
=
1
-
\frac{
-\sum_{g=1}^{K}
p_{ug}\log(p_{ug}+\delta)
}{
\log K
}.
\end{equation}

The mean specialization is:

\begin{equation}
S_{\mathrm{spec}}
=
\frac{1}{n}
\sum_{u=1}^{n}
\operatorname{Spec}_u.
\end{equation}

A value close to 1 indicates that a neuron concentrates its activity in one group, while a value close to 0 indicates that its activity is distributed approximately uniformly across groups.

\paragraph{Purity.}
Purity uses the same group-conditioned distribution but retains only the largest group probability:

\begin{equation}
\operatorname{Purity}_u
=
\max_g p_{ug}.
\end{equation}

The average purity is:

\begin{equation}
P_{\mathrm{avg}}
=
\frac{1}{n}
\sum_{u=1}^{n}
\operatorname{Purity}_u.
\end{equation}

Purity provides a simpler measure of group preference, although it contains less information than the entropy-based specialization score.

\paragraph{Centroid separation.}
Centroid separation measures how well the groups separate in hidden space. For group \(g\), its centroid is:

\begin{equation}
c_g
=
\frac{1}{|\mathcal{I}_g|}
\sum_{i \in \mathcal{I}_g}
H_i.
\end{equation}

The within-group squared distance for sample \(i \in \mathcal{I}_g\) is:

\begin{equation}
d_{\mathrm{within}}(i,g)
=
\|H_i-c_g\|_2^2.
\end{equation}

The squared distance between group centroids is:

\begin{equation}
d_{\mathrm{between}}(g,h)
=
\|c_g-c_h\|_2^2.
\end{equation}

The centroid-separation score is:

\begin{equation}
S_{\mathrm{centroid}}
=
\frac{
\operatorname{mean}_{g<h}
d_{\mathrm{between}}(g,h)
}{
\max\!\left(
\operatorname{mean}_{g,\,i\in\mathcal{I}_g}
d_{\mathrm{within}}(i,g),
\delta
\right)
}.
\end{equation}

High values indicate compact groups whose centroids are well separated.

\paragraph{Within-group and between-group Jaccard similarity.}
Within-group Jaccard similarity measures how similar the sets of active neurons are for samples from the same group. Let us define the active support of sample \(i\) as \(A_i=\{u:|H_{iu}|>\varepsilon\}\). For two samples:

\begin{equation}
J(i,j)
=
\frac{
|A_i \cap A_j|
}{
|A_i \cup A_j|
}.
\end{equation}

If both supports are empty, \(J(i,j)\) is defined as 1. The within-group score \(J_{\mathrm{within}}\) is the mean \(J(i,j)\) over 8192 sampled pairs from the same group. High values indicate that samples from the same group tend to activate similar subsets of hidden neurons.

Between-group Jaccard similarity uses the same definition for 8192 sampled pairs drawn from different groups. Lower values indicate that different groups rely on more distinct hidden-neuron supports. The Jaccard gap is:

\begin{equation}
J_{\mathrm{gap}}
=
J_{\mathrm{within}}
-
J_{\mathrm{between}}.
\end{equation}

A large positive gap indicates that active-neuron support is more similar within groups than between groups.

\paragraph{Activation sparsity.}
Activation sparsity is also reported on the task's training set:

\begin{equation}
S_{\mathrm{act}}
=
\frac{1}{N_{\mathrm{tr}}n}
\sum_{i=1}^{N_{\mathrm{tr}}}
\sum_{u=1}^{n}
\mathbf{1}\!\left[H_{iu}\leq\varepsilon\right].
\end{equation}

This quantity has the same algebraic form as population sparsity; the distinction in the experiments is the evaluation set. The grouped structure indicators are evaluated on the combined training and test samples, whereas activation sparsity is reported on the training samples.

\paragraph{Structural sparsity.}
Structural sparsity measures the fraction of inactive connections in the effective first hidden layer. When structural plasticity is active:

\begin{equation}
S_{\mathrm{struct}}
=
1
-
\frac{1}{nd}
\sum_{i=1}^{n}
\sum_{j=1}^{d}
M_{ij}.
\end{equation}

When structural plasticity is disabled, the effective layer is dense and \(S_{\mathrm{struct}}=0\). This metric describes connectivity sparsity rather than activation sparsity.

\paragraph{Active subnetwork size.}
Finally, active subnetwork size is reported for sparse parity as a functional measure of how many hidden neurons are needed to reproduce the decisions of the full model. Hidden neurons are ordered as \(u_1,\ldots,u_n\) by decreasing norm of their effective incoming weights. For sample \(i\), let us define the full output score:

\begin{equation}
\hat{y}_i
=
\sum_{k=1}^{n}
H_{i,u_k}
(W_y)_{1,u_k}.
\end{equation}

The score produced by the first \(m\) neurons is:

\begin{equation}
\hat{y}_i^{(m)}
=
\sum_{k=1}^{m}
H_{i,u_k}
(W_y)_{1,u_k}.
\end{equation}

The active subnetwork size is the smallest \(m\) such that:

\begin{equation}
\operatorname{sign}\!\left(\hat{y}_i^{(m)}\right)
=
\operatorname{sign}(\hat{y}_i)
\end{equation}
for every sample in the support set (e.g., the training set). A small value means that a compact subset of high-utility hidden neurons reproduces the full model's decisions, while a large value indicates that the decision is distributed across more neurons. This metric is used for sparse parity and is not reported for noisy XOR.

\subsection{Experimental Protocol}

Both tasks use a BioNN model with one hidden layer of 1000 neurons and a scalar output. All ablation configurations instantiate the same underlying architecture and modulation heads; disabling a mechanism replaces its contribution with its neutral behavior rather than changing the architecture. This keeps model capacity and initialization structure comparable across ablations. The same set of random seeds is used for all configurations of a task, so corresponding ablations are evaluated under matched initialization and data-generation conditions.

Sparse parity uses 1000 training examples and 100 test examples, with minibatches of 32. Training uses SGD with a learning rate of 0.1, weight decay of 0.01, hinge loss, and 300 epochs. Structural density is $\rho_m=0.35$, and the structural mask is updated every 10 epochs when structural plasticity is enabled. The training set is regenerated for each independent run, while the same held-out test set is used across runs.

Noisy XOR uses 200 training examples and 200 test examples. Training uses full-batch SGD with a learning rate of 0.05, weight decay of $10^{-4}$, logistic loss, and 5000 optimization steps. The experiments are conducted with structural densities of $\rho_m=1$ (thus with structural plasticity disabled) and $\rho_m=0.35$. Both the training and test samples are independently generated for each run.

For each experimental configuration (combination of mechanisms), 20 independent runs were carried out.

The use of hinge loss is important in this setting because reaching 100\% training accuracy does not terminate the learning signal. Correctly classified examples that are still too close to the decision boundary continue to contribute to the loss, so training can keep refining the internal representation.

Some additional tests were conducted with more recent optimizers, such as Adam. Interestingly, they produced substantially different results for the grokking experiments, with generally worse performance and different generalization dynamics, including cases in which generalization appeared to be delayed or was not observed within the evaluation period. Further experiments would be needed to draw a clear conclusion about the effect of the optimizer. For the present study, SGD was chosen because it was also used in the original paper or code associated with the two grokking problems considered here.

\section{Results and Discussion}

\subsection{Ablation Results}

Tables 2--4 report the experimental results for the parity and XOR tasks, respectively. Each row corresponds to one experiment configuration. The columns present the indicator variables, training and test measurements, sparsity measures, representation statistics, and overlap measures.

The first columns identify the experimental configuration. IG, GM, TM, LI, HO, SP, and AD denote input gating, gain modulation, threshold modulation, lateral inhibition, homeostasis, structural plasticity, and activation decorrelation, respectively; a value of 1 indicates that the corresponding mechanism is enabled, while 0 indicates that it is disabled. The experiment names provide a compact description of the same configuration: ``all on'' and ``all off'' denote the two extreme cases; ``one off $X$'' means that all mechanisms are enabled except $X$; ``two off $X Y$'' means that all mechanisms are enabled except $X$ and $Y$; ``one on $X$'' means that only $X$ is enabled; and ``two on $X Y$'' means that only $X$ and $Y$ are enabled. The columns ``Train 0.9'' and ``Test 0.9'' report the first training epoch at which training and test accuracy, respectively, reach 0.9, while ``Train 1'' and ``Test 1'' report the first epoch at which perfect accuracy is reached. The values in the tables are averages over the runs that reach that threshold. A value of 0 indicates that the corresponding accuracy threshold was not reached within the experimental horizon. The remaining columns report the structural and representation indicators defined in the previous section.

\begin{landscape}

\begingroup
\tiny
\setlength{\tabcolsep}{1.2pt}
\renewcommand{\arraystretch}{1.05}
\setlength{\LTleft}{0pt plus 1fill}
\setlength{\LTright}{0pt plus 1fill}

\begin{longtable}{@{}>{\raggedright\arraybackslash}p{4.35cm}*{26}{c}@{}}
\caption{Experimental results for the parity task.}
\label{tab:parity-results}\\

\toprule
\textbf{Experiment} & \rotatebox{90}{\textbf{IG}} & \rotatebox{90}{\textbf{GM}} & \rotatebox{90}{\textbf{TM}} & \rotatebox{90}{\textbf{LI}} & \rotatebox{90}{\textbf{HO}} & \rotatebox{90}{\textbf{SP}} & \rotatebox{90}{\textbf{AD}} & \rotatebox{90}{\textbf{Train 0.9}} & \rotatebox{90}{\textbf{Test 0.9}} & \rotatebox{90}{\textbf{Train 1}} & \rotatebox{90}{\textbf{Test 1}} & \rotatebox{90}{\textbf{Lifetime Sparsity}} & \rotatebox{90}{\textbf{Sample Hoyer}} & \rotatebox{90}{\textbf{Neuron Hoyer}} & \rotatebox{90}{\textbf{Dead Neuron}} & \rotatebox{90}{\textbf{Mean Abs Corr}} & \rotatebox{90}{\textbf{Effective Rank }} & \rotatebox{90}{\textbf{Specialization }} & \rotatebox{90}{\textbf{Purity }} & \rotatebox{90}{\textbf{Centroid Sep }} & \rotatebox{90}{\textbf{Within Jaccard }} & \rotatebox{90}{\textbf{Between Jaccard }} & \rotatebox{90}{\textbf{Jaccard Gap }} & \rotatebox{90}{\textbf{Activation Sparsity}} & \rotatebox{90}{\textbf{Structural Sparsity}} & \rotatebox{90}{\textbf{Active Subnetwork Size}} \\
\midrule
\endfirsthead

\caption[]{Experimental results for the parity task (continued).}\\
\toprule
\textbf{Experiment} & \rotatebox{90}{\textbf{IG}} & \rotatebox{90}{\textbf{GM}} & \rotatebox{90}{\textbf{TM}} & \rotatebox{90}{\textbf{LI}} & \rotatebox{90}{\textbf{HO}} & \rotatebox{90}{\textbf{SP}} & \rotatebox{90}{\textbf{AD}} & \rotatebox{90}{\textbf{Train 0.9}} & \rotatebox{90}{\textbf{Test 0.9}} & \rotatebox{90}{\textbf{Train 1}} & \rotatebox{90}{\textbf{Test 1}} & \rotatebox{90}{\textbf{Lifetime Sparsity }} & \rotatebox{90}{\textbf{Sample Hoyer}} & \rotatebox{90}{\textbf{Neuron Hoyer }} & \rotatebox{90}{\textbf{Dead Neuron}} & \rotatebox{90}{\textbf{Mean Abs Corr }} & \rotatebox{90}{\textbf{Effective Rank }} & \rotatebox{90}{\textbf{Specialization }} & \rotatebox{90}{\textbf{Purity }} & \rotatebox{90}{\textbf{Centroid Sep }} & \rotatebox{90}{\textbf{Within Jaccard }} & \rotatebox{90}{\textbf{Between Jaccard }} & \rotatebox{90}{\textbf{Jaccard Gap }} & \rotatebox{90}{\textbf{Activation Sparsity}} & \rotatebox{90}{\textbf{Structural Sparsity}} & \rotatebox{90}{\textbf{Active Subnetwork Size}} \\
\midrule
\endhead

\midrule
\multicolumn{27}{r}{Continued on next page}\\
\endfoot

\bottomrule
\endlastfoot

\texttt{two off   inhibition   decorrelation} & 1 & 1 & 1 & 0 & 1 & 1 & 0 & 12 & 14 & 16 & 17 & 0.863 & 0.999 & 0.855 & 0.833 & 0.129 & 0.007 & 0.024 & 0.096 & 0.671 & 0.943 & 0.932 & 0.011 & 0.863 & 0.65 & 7 \\
\texttt{two off   gain   threshold} & 1 & 0 & 0 & 1 & 1 & 1 & 1 & 13 & 15 & 16 & 17 & 0.862 & 0.991 & 0.821 & 0.559 & 0.066 & 0.012 & 0.062 & 0.268 & 0.472 & 0.495 & 0.488 & 0.007 & 0.862 & 0.65 & 9 \\
\texttt{one off   threshold} & 1 & 1 & 0 & 1 & 1 & 1 & 1 & 12 & 14 & 16 & 18 & 0.869 & 0.995 & 0.829 & 0.585 & 0.068 & 0.010 & 0.044 & 0.243 & 0.518 & 0.542 & 0.536 & 0.006 & 0.869 & 0.65 & 8 \\
\texttt{two off   threshold   decorrelation} & 1 & 1 & 0 & 1 & 1 & 1 & 0 & 12 & 14 & 16 & 18 & 0.927 & 0.999 & 0.909 & 0.830 & 0.089 & 0.007 & 0.030 & 0.106 & 0.670 & 0.916 & 0.896 & 0.020 & 0.927 & 0.65 & 7 \\
\texttt{one off   decorrelation} & 1 & 1 & 1 & 1 & 1 & 1 & 0 & 12 & 14 & 16 & 18 & 0.934 & 0.999 & 0.924 & 0.841 & 0.072 & 0.007 & 0.053 & 0.111 & 0.670 & 0.914 & 0.891 & 0.023 & 0.934 & 0.65 & 7 \\
\texttt{one off   inhibition} & 1 & 1 & 1 & 0 & 1 & 1 & 1 & 12 & 14 & 16 & 18 & 0.899 & 0.998 & 0.882 & 0.811 & 0.082 & 0.009 & 0.023 & 0.109 & 0.666 & 0.708 & 0.701 & 0.007 & 0.899 & 0.65 & 7 \\
\texttt{two off   threshold   inhibition} & 1 & 1 & 0 & 0 & 1 & 1 & 1 & 12 & 14 & 16 & 18 & 0.846 & 0.997 & 0.829 & 0.751 & 0.091 & 0.008 & 0.024 & 0.140 & 0.665 & 0.737 & 0.732 & 0.005 & 0.846 & 0.65 & 8 \\
\texttt{one off   gain} & 1 & 0 & 1 & 1 & 1 & 1 & 1 & 13 & 14 & 16 & 18 & 0.845 & 0.993 & 0.804 & 0.547 & 0.074 & 0.010 & 0.052 & 0.269 & 0.494 & 0.552 & 0.545 & 0.007 & 0.845 & 0.65 & 9 \\
\texttt{two off   gain   decorrelation} & 1 & 0 & 1 & 1 & 1 & 1 & 0 & 13 & 14 & 16 & 18 & 0.862 & 0.994 & 0.852 & 0.808 & 0.365 & 0.007 & 0.041 & 0.120 & 0.669 & 0.925 & 0.910 & 0.015 & 0.862 & 0.65 & 7 \\
\texttt{two off   gain   inhibition} & 1 & 0 & 1 & 0 & 1 & 1 & 1 & 13 & 14 & 16 & 18 & 0.898 & 0.994 & 0.882 & 0.807 & 0.082 & 0.009 & 0.033 & 0.116 & 0.638 & 0.698 & 0.688 & 0.010 & 0.898 & 0.65 & 8 \\
\texttt{all on} & 1 & 1 & 1 & 1 & 1 & 1 & 1 & 12 & 15 & 16 & 18 & 0.858 & 0.993 & 0.812 & 0.541 & 0.071 & 0.011 & 0.035 & 0.262 & 0.499 & 0.526 & 0.522 & 0.004 & 0.858 & 0.65 & 10 \\
\texttt{two off   input gate   inhibition} & 0 & 1 & 1 & 0 & 1 & 1 & 1 & 13 & 15 & 17 & 19 & 0.882 & 0.997 & 0.865 & 0.782 & 0.084 & 0.009 & 0.027 & 0.127 & 0.628 & 0.693 & 0.687 & 0.006 & 0.882 & 0.65 & 7 \\
\texttt{one off   input gate} & 0 & 1 & 1 & 1 & 1 & 1 & 1 & 13 & 15 & 17 & 20 & 0.873 & 0.992 & 0.821 & 0.474 & 0.068 & 0.011 & 0.051 & 0.311 & 0.399 & 0.440 & 0.436 & 0.004 & 0.873 & 0.65 & 11 \\
\texttt{two off   input gate   decorrelation} & 0 & 1 & 1 & 1 & 1 & 1 & 0 & 13 & 15 & 17 & 20 & 0.937 & 0.999 & 0.920 & 0.825 & 0.065 & 0.007 & 0.039 & 0.115 & 0.671 & 0.848 & 0.828 & 0.021 & 0.937 & 0.65 & 7 \\
\texttt{two off   input gate   threshold} & 0 & 1 & 0 & 1 & 1 & 1 & 1 & 13 & 15 & 18 & 20 & 0.876 & 0.997 & 0.851 & 0.705 & 0.080 & 0.009 & 0.032 & 0.172 & 0.566 & 0.627 & 0.622 & 0.004 & 0.876 & 0.65 & 7 \\
\texttt{two off   input gate   gain} & 0 & 0 & 1 & 1 & 1 & 1 & 1 & 13 & 16 & 18 & 20 & 0.868 & 0.993 & 0.838 & 0.660 & 0.068 & 0.011 & 0.044 & 0.203 & 0.541 & 0.602 & 0.594 & 0.008 & 0.868 & 0.65 & 8 \\
\texttt{two on   homeostasis   structural} & 0 & 0 & 0 & 0 & 1 & 1 & 0 & 14 & 17 & 19 & 22 & 0.872 & 0.997 & 0.857 & 0.804 & 0.106 & 0.007 & 0.038 & 0.121 & 0.668 & 0.857 & 0.845 & 0.011 & 0.871 & 0.65 & 7 \\
\texttt{two off   input gate   homeostasis} & 0 & 1 & 1 & 1 & 0 & 1 & 1 & 18 & 25 & 25 & 31 & 0.762 & 0.909 & 0.730 & 0.366 & 0.088 & 0.013 & 0.116 & 0.402 & 0.336 & 0.476 & 0.468 & 0.008 & 0.762 & 0.65 & 8 \\
\texttt{one off   homeostasis} & 1 & 1 & 1 & 1 & 0 & 1 & 1 & 17 & 25 & 26 & 35 & 0.833 & 0.920 & 0.795 & 0.465 & 0.118 & 0.011 & 0.070 & 0.328 & 0.793 & 0.392 & 0.377 & 0.015 & 0.833 & 0.65 & 8 \\
\texttt{two off   homeostasis   decorrelation} & 1 & 1 & 1 & 1 & 0 & 1 & 0 & 17 & 25 & 25 & 35 & 0.749 & 0.920 & 0.656 & 0.268 & 0.260 & 0.007 & 0.271 & 0.557 & 0.896 & 0.426 & 0.344 & 0.082 & 0.749 & 0.65 & 7 \\
\texttt{two off   gain   homeostasis} & 1 & 0 & 1 & 1 & 0 & 1 & 1 & 18 & 29 & 28 & 36 & 0.846 & 0.895 & 0.772 & 0.294 & 0.109 & 0.013 & 0.159 & 0.475 & 0.579 & 0.273 & 0.218 & 0.055 & 0.846 & 0.65 & 24 \\
\texttt{two off   threshold   homeostasis} & 1 & 1 & 0 & 1 & 0 & 1 & 1 & 18 & 26 & 28 & 38 & 0.786 & 0.934 & 0.736 & 0.388 & 0.112 & 0.014 & 0.056 & 0.360 & 0.598 & 0.340 & 0.333 & 0.008 & 0.786 & 0.65 & 6 \\
\texttt{two on   gain   structural} & 0 & 1 & 0 & 0 & 0 & 1 & 0 & 21 & 31 & 31 & 39 & 0.516 & 0.882 & 0.335 & 0.000 & 0.126 & 0.011 & 0.016 & 0.541 & 0.756 & 0.480 & 0.474 & 0.006 & 0.516 & 0.65 & 6 \\
\texttt{two off   inhibition   homeostasis} & 1 & 1 & 1 & 0 & 0 & 1 & 1 & 19 & 29 & 29 & 39 & 0.769 & 0.911 & 0.746 & 0.328 & 0.068 & 0.019 & 0.190 & 0.456 & 0.314 & 0.435 & 0.418 & 0.017 & 0.769 & 0.65 & 14 \\
\texttt{two on   inhibition   structural} & 0 & 0 & 0 & 1 & 0 & 1 & 0 & 20 & 31 & 31 & 39 & 0.684 & 0.958 & 0.552 & 0.014 & 0.143 & 0.010 & 0.201 & 0.675 & 0.366 & 0.258 & 0.232 & 0.026 & 0.685 & 0.65 & 6 \\
\texttt{two on   threshold   structural} & 0 & 0 & 1 & 0 & 0 & 1 & 0 & 21 & 33 & 32 & 42 & 0.671 & 0.955 & 0.440 & 0.000 & 0.112 & 0.008 & 0.030 & 0.571 & 0.409 & 0.280 & 0.269 & 0.011 & 0.671 & 0.65 & 6 \\
\texttt{one on   structural} & 0 & 0 & 0 & 0 & 0 & 1 & 0 & 22 & 35 & 33 & 43 & 0.652 & 0.958 & 0.443 & 0.000 & 0.109 & 0.009 & 0.025 & 0.568 & 0.319 & 0.267 & 0.261 & 0.007 & 0.652 & 0.65 & 5 \\
\texttt{two on   structural   decorrelation} & 0 & 0 & 0 & 0 & 0 & 1 & 1 & 22 & 37 & 34 & 45 & 0.879 & 0.951 & 0.838 & 0.377 & 0.037 & 0.014 & 0.177 & 0.434 & 0.135 & 0.270 & 0.262 & 0.008 & 0.880 & 0.65 & 10 \\
\texttt{two off   structural   decorrelation} & 1 & 1 & 1 & 1 & 1 & 0 & 0 & 8 & 42 & 18 & 53 & 0.874 & 0.997 & 0.839 & 0.676 & 0.155 & 0.007 & 0.166 & 0.253 & 0.670 & 0.703 & 0.645 & 0.058 & 0.874 & 0.00 & 7 \\
\texttt{two off   inhibition   structural} & 1 & 1 & 1 & 0 & 1 & 0 & 1 & 8 & 42 & 18 & 53 & 0.858 & 0.993 & 0.813 & 0.500 & 0.055 & 0.011 & 0.115 & 0.326 & 0.400 & 0.483 & 0.471 & 0.012 & 0.858 & 0.00 & 11 \\
\texttt{one off   structural} & 1 & 1 & 1 & 1 & 1 & 0 & 1 & 8 & 42 & 18 & 53 & 0.844 & 0.993 & 0.768 & 0.228 & 0.051 & 0.011 & 0.119 & 0.481 & 0.367 & 0.299 & 0.291 & 0.008 & 0.844 & 0.00 & 11 \\
\texttt{two on   input gate   structural} & 1 & 0 & 0 & 0 & 0 & 1 & 0 & 22 & 39 & 36 & 54 & 0.571 & 0.814 & 0.409 & 0.000 & 0.115 & 0.067 & 0.046 & 0.576 & 0.399 & 0.355 & 0.344 & 0.010 & 0.571 & 0.65 & 36 \\
\texttt{two off   gain   structural} & 1 & 0 & 1 & 1 & 1 & 0 & 1 & 8 & 43 & 18 & 54 & 0.851 & 0.986 & 0.755 & 0.126 & 0.052 & 0.012 & 0.209 & 0.584 & 0.281 & 0.220 & 0.200 & 0.020 & 0.851 & 0.00 & 11 \\
\texttt{two off   threshold   structural} & 1 & 1 & 0 & 1 & 1 & 0 & 1 & 8 & 43 & 19 & 54 & 0.850 & 0.991 & 0.772 & 0.228 & 0.051 & 0.013 & 0.125 & 0.484 & 0.418 & 0.291 & 0.282 & 0.010 & 0.850 & 0.00 & 12 \\
\texttt{two on   input gate   homeostasis} & 1 & 0 & 0 & 0 & 1 & 0 & 0 & 8 & 44 & 19 & 57 & 0.884 & 0.990 & 0.817 & 0.517 & 0.133 & 0.008 & 0.326 & 0.416 & 0.668 & 0.475 & 0.389 & 0.086 & 0.884 & 0.00 & 7 \\
\texttt{two off   input gate   structural} & 0 & 1 & 1 & 1 & 1 & 0 & 1 & 8 & 49 & 20 & 62 & 0.831 & 0.990 & 0.746 & 0.139 & 0.051 & 0.012 & 0.108 & 0.525 & 0.251 & 0.260 & 0.254 & 0.005 & 0.831 & 0.00 & 11 \\
\texttt{two on   gain   homeostasis} & 0 & 1 & 0 & 0 & 1 & 0 & 0 & 9 & 50 & 20 & 66 & 0.868 & 0.997 & 0.829 & 0.607 & 0.078 & 0.008 & 0.132 & 0.284 & 0.669 & 0.562 & 0.544 & 0.018 & 0.868 & 0.00 & 7 \\
\texttt{two on   threshold   homeostasis} & 0 & 0 & 1 & 0 & 1 & 0 & 0 & 9 & 53 & 21 & 67 & 0.867 & 0.991 & 0.800 & 0.418 & 0.071 & 0.011 & 0.219 & 0.444 & 0.668 & 0.366 & 0.340 & 0.025 & 0.867 & 0.00 & 8 \\
\texttt{two on   inhibition   homeostasis} & 0 & 0 & 0 & 1 & 1 & 0 & 0 & 9 & 53 & 21 & 68 & 0.850 & 0.990 & 0.795 & 0.481 & 0.090 & 0.011 & 0.194 & 0.393 & 0.667 & 0.503 & 0.473 & 0.030 & 0.850 & 0.00 & 8 \\
\texttt{one on   homeostasis} & 0 & 0 & 0 & 0 & 1 & 0 & 0 & 9 & 54 & 21 & 69 & 0.865 & 0.991 & 0.797 & 0.399 & 0.069 & 0.011 & 0.224 & 0.459 & 0.668 & 0.367 & 0.343 & 0.024 & 0.864 & 0.00 & 7 \\
\texttt{two on   homeostasis   decorrelation} & 0 & 0 & 0 & 0 & 1 & 0 & 1 & 9 & 53 & 21 & 70 & 0.857 & 0.987 & 0.790 & 0.346 & 0.051 & 0.011 & 0.169 & 0.451 & 0.219 & 0.352 & 0.336 & 0.015 & 0.857 & 0.00 & 10 \\
\texttt{two off   homeostasis   structural} & 1 & 1 & 1 & 1 & 0 & 0 & 1 & 12 & 96 & 42 & 116 & 0.699 & 0.796 & 0.613 & 0.182 & 0.194 & 0.019 & 0.229 & 0.589 & 0.869 & 0.358 & 0.260 & 0.098 & 0.699 & 0.00 & 102 \\
\texttt{two on   threshold   inhibition} & 0 & 0 & 1 & 1 & 0 & 0 & 0 & 13 & 101 & 64 & 116 & 0.589 & 0.872 & 0.501 & 0.001 & 0.145 & 0.052 & 0.081 & 0.610 & 0.419 & 0.296 & 0.289 & 0.007 & 0.590 & 0.00 & 7 \\
\texttt{two on   gain   inhibition} & 0 & 1 & 0 & 1 & 0 & 0 & 0 & 14 & 104 & 58 & 119 & 0.591 & 0.821 & 0.490 & 0.006 & 0.300 & 0.030 & 0.113 & 0.642 & 0.440 & 0.341 & 0.297 & 0.044 & 0.592 & 0.00 & 6 \\
\texttt{two on   gain   threshold} & 0 & 1 & 1 & 0 & 0 & 0 & 0 & 14 & 105 & 63 & 120 & 0.571 & 0.875 & 0.468 & 0.000 & 0.157 & 0.027 & 0.080 & 0.612 & 0.538 & 0.363 & 0.351 & 0.012 & 0.571 & 0.00 & 6 \\
\texttt{two on   input gate   threshold} & 1 & 0 & 1 & 0 & 0 & 0 & 0 & 13 & 104 & 52 & 121 & 0.539 & 0.637 & 0.414 & 0.000 & 0.310 & 0.022 & 0.299 & 0.754 & 1.178 & 0.393 & 0.277 & 0.116 & 0.539 & 0.00 & 127 \\
\texttt{one on   inhibition} & 0 & 0 & 0 & 1 & 0 & 0 & 0 & 14 & 109 & 66 & 123 & 0.591 & 0.861 & 0.504 & 0.000 & 0.139 & 0.069 & 0.078 & 0.618 & 0.408 & 0.293 & 0.288 & 0.005 & 0.591 & 0.00 & 16 \\
\texttt{two on   inhibition   decorrelation} & 0 & 0 & 0 & 1 & 0 & 0 & 1 & 13 & 107 & 73 & 123 & 0.651 & 0.880 & 0.566 & 0.005 & 0.087 & 0.054 & 0.078 & 0.610 & 0.234 & 0.278 & 0.274 & 0.004 & 0.652 & 0.00 & 16 \\
\texttt{one on   threshold} & 0 & 0 & 1 & 0 & 0 & 0 & 0 & 14 & 119 & 78 & 133 & 0.554 & 0.873 & 0.472 & 0.000 & 0.146 & 0.060 & 0.061 & 0.594 & 0.411 & 0.312 & 0.308 & 0.005 & 0.555 & 0.00 & 8 \\
\texttt{two on   threshold   decorrelation} & 0 & 0 & 1 & 0 & 0 & 0 & 1 & 14 & 119 & 64 & 133 & 0.667 & 0.926 & 0.580 & 0.002 & 0.074 & 0.021 & 0.097 & 0.619 & 0.065 & 0.262 & 0.259 & 0.004 & 0.668 & 0.00 & 11 \\
\texttt{two on   gain   decorrelation} & 0 & 1 & 0 & 0 & 0 & 0 & 1 & 15 & 122 & 79 & 136 & 0.607 & 0.874 & 0.548 & 0.005 & 0.086 & 0.043 & 0.089 & 0.597 & 0.205 & 0.390 & 0.384 & 0.006 & 0.608 & 0.00 & 9 \\
\texttt{all off} & 0 & 0 & 0 & 0 & 0 & 0 & 0 & 15 & 120 & 70 & 137 & 0.554 & 0.857 & 0.472 & 0.000 & 0.135 & 0.066 & 0.055 & 0.596 & 0.365 & 0.309 & 0.303 & 0.006 & 0.555 & 0.00 & 10 \\
\texttt{one on   gain} & 0 & 1 & 0 & 0 & 0 & 0 & 0 & 15 & 122 & 71 & 138 & 0.563 & 0.832 & 0.468 & 0.000 & 0.114 & 0.071 & 0.073 & 0.604 & 0.618 & 0.382 & 0.373 & 0.009 & 0.564 & 0.00 & 10 \\
\texttt{one on   decorrelation} & 0 & 0 & 0 & 0 & 0 & 0 & 1 & 15 & 129 & 76 & 146 & 0.620 & 0.864 & 0.541 & 0.001 & 0.087 & 0.065 & 0.065 & 0.597 & 0.219 & 0.294 & 0.291 & 0.003 & 0.621 & 0.00 & 20 \\
\texttt{two on   input gate   inhibition} & 1 & 0 & 0 & 1 & 0 & 0 & 0 & 13 & 145 & 41 & 176 & 0.582 & 0.675 & 0.501 & 0.000 & 0.158 & 0.231 & 0.099 & 0.651 & 0.290 & 0.289 & 0.275 & 0.014 & 0.583 & 0.00 & 82 \\
\texttt{two on   input gate   decorrelation} & 1 & 0 & 0 & 0 & 0 & 0 & 1 & 13 & 132 & 46 & 200 & 0.570 & 0.648 & 0.499 & 0.000 & 0.129 & 0.228 & 0.101 & 0.648 & 0.293 & 0.318 & 0.294 & 0.024 & 0.570 & 0.00 & 143 \\
\texttt{one on   input gate} & 1 & 0 & 0 & 0 & 0 & 0 & 0 & 14 & 140 & 53 & 201 & 0.540 & 0.637 & 0.468 & 0.000 & 0.161 & 0.231 & 0.102 & 0.653 & 0.330 & 0.328 & 0.299 & 0.029 & 0.541 & 0.00 & 115 \\
\texttt{two on   input gate   gain} & 1 & 1 & 0 & 0 & 0 & 0 & 0 & 14 & 171 & 42 & 215 & 0.552 & 0.751 & 0.472 & 0.000 & 0.199 & 0.108 & 0.108 & 0.630 & 1.074 & 0.385 & 0.357 & 0.028 & 0.553 & 0.00 & 45 \\

\end{longtable}
\endgroup

\begingroup
\tiny
\setlength{\tabcolsep}{1.2pt}
\renewcommand{\arraystretch}{1.05}
\setlength{\LTleft}{0pt plus 1fill}
\setlength{\LTright}{0pt plus 1fill}

\begin{longtable}{@{}>{\raggedright\arraybackslash}p{4.35cm}*{24}{c}@{}}
\caption{Experimental results for the XOR task.}
\label{tab:xor-results}\\

\toprule
\textbf{Experiment} & \rotatebox{90}{\textbf{IG}} & \rotatebox{90}{\textbf{GM}} & \rotatebox{90}{\textbf{TM}} & \rotatebox{90}{\textbf{LI}} & \rotatebox{90}{\textbf{HO}} & \rotatebox{90}{\textbf{SP}} & \rotatebox{90}{\textbf{AD}} & \rotatebox{90}{\textbf{Train 0.9}} & \rotatebox{90}{\textbf{Test 0.9}} & \rotatebox{90}{\textbf{Train 1}} & \rotatebox{90}{\textbf{Test 1}} & \rotatebox{90}{\textbf{Lifetime Sparsity }} & \rotatebox{90}{\textbf{Sample Hoyer }} & \rotatebox{90}{\textbf{Neuron Hoyer }} & \rotatebox{90}{\textbf{Dead Neuron }} & \rotatebox{90}{\textbf{Mean Abs. Corr. }} & \rotatebox{90}{\textbf{Effective Rank }} & \rotatebox{90}{\textbf{Specialization }} & \rotatebox{90}{\textbf{Purity }} & \rotatebox{90}{\textbf{Centroid Sep. }} & \rotatebox{90}{\textbf{Within Jaccard }} & \rotatebox{90}{\textbf{Between Jaccard }} & \rotatebox{90}{\textbf{Jaccard Gap }} & \rotatebox{90}{\textbf{Activation Sparsity}} \\
\midrule
\endfirsthead

\caption[]{Experimental results for the XOR task (continued).}\\
\toprule
\textbf{Experiment} & \rotatebox{90}{\textbf{IG}} & \rotatebox{90}{\textbf{GM}} & \rotatebox{90}{\textbf{TM}} & \rotatebox{90}{\textbf{LI}} & \rotatebox{90}{\textbf{HO}} & \rotatebox{90}{\textbf{SP}} & \rotatebox{90}{\textbf{AD}} & \rotatebox{90}{\textbf{Train 0.9}} & \rotatebox{90}{\textbf{Test 0.9}} & \rotatebox{90}{\textbf{Train 1}} & \rotatebox{90}{\textbf{Test 1}} & \rotatebox{90}{\textbf{Lifetime Sparsity }} & \rotatebox{90}{\textbf{Sample Hoyer }} & \rotatebox{90}{\textbf{Neuron Hoyer }} & \rotatebox{90}{\textbf{Dead Neuron }} & \rotatebox{90}{\textbf{Mean Abs. Corr. }} & \rotatebox{90}{\textbf{Effective Rank }} & \rotatebox{90}{\textbf{Specialization }} & \rotatebox{90}{\textbf{Purity }} & \rotatebox{90}{\textbf{Centroid Sep. }} & \rotatebox{90}{\textbf{Within Jaccard }} & \rotatebox{90}{\textbf{Between Jaccard }} & \rotatebox{90}{\textbf{Jaccard Gap }} & \rotatebox{90}{\textbf{Activation Sparsity}} \\
\midrule
\endhead

\midrule
\multicolumn{25}{r}{Continued on next page}\\
\endfoot

\bottomrule
\endlastfoot

\texttt{all on} & 1 & 1 & 1 & 1 & 1 & 1 & 1 & 99 & 1691 & 219 & 2310 & 0.832 & 0.823 & 0.745 & 0.000 & 0.139 & 0.197 & 0.720 & 0.858 & 28.779 & 0.392 & 0.170 & 0.222 & 0.818 \\
\texttt{two off   threshold   inhibition} & 1 & 1 & 0 & 0 & 1 & 1 & 1 & 96 & 1692 & 213 & 2463 & 0.845 & 0.828 & 0.756 & 0.000 & 0.139 & 0.185 & 0.745 & 0.871 & 29.973 & 0.401 & 0.173 & 0.227 & 0.820 \\
\texttt{one off   inhibition} & 1 & 1 & 1 & 0 & 1 & 1 & 1 & 96 & 1698 & 215 & 2470 & 0.845 & 0.827 & 0.756 & 0.000 & 0.139 & 0.185 & 0.745 & 0.871 & 29.900 & 0.398 & 0.173 & 0.226 & 0.820 \\
\texttt{two off   inhibition   structural} & 1 & 1 & 1 & 0 & 1 & 0 & 1 & 96 & 1698 & 215 & 2470 & 0.845 & 0.827 & 0.756 & 0.000 & 0.139 & 0.185 & 0.745 & 0.871 & 29.900 & 0.398 & 0.173 & 0.226 & 0.820 \\
\texttt{one off   structural} & 1 & 1 & 1 & 1 & 1 & 0 & 1 & 97 & 1703 & 215 & 2475 & 0.836 & 0.828 & 0.747 & 0.000 & 0.138 & 0.191 & 0.725 & 0.858 & 29.967 & 0.385 & 0.166 & 0.219 & 0.821 \\
\texttt{two off   inhibition   decorrelation} & 1 & 1 & 1 & 0 & 1 & 1 & 0 & 97 & 1693 & 214 & 2481 & 0.844 & 0.827 & 0.756 & 0.000 & 0.141 & 0.185 & 0.754 & 0.877 & 29.934 & 0.404 & 0.175 & 0.229 & 0.819 \\
\texttt{one off   decorrelation} & 1 & 1 & 1 & 1 & 1 & 1 & 0 & 96 & 1696 & 215 & 2484 & 0.836 & 0.828 & 0.747 & 0.000 & 0.140 & 0.190 & 0.734 & 0.865 & 29.976 & 0.387 & 0.166 & 0.222 & 0.821 \\
\texttt{two off   structural   decorrelation} & 1 & 1 & 1 & 1 & 1 & 0 & 0 & 96 & 1696 & 215 & 2484 & 0.836 & 0.828 & 0.747 & 0.000 & 0.140 & 0.190 & 0.734 & 0.865 & 29.976 & 0.387 & 0.166 & 0.222 & 0.821 \\
\texttt{two off   input gate   inhibition} & 0 & 1 & 1 & 0 & 1 & 1 & 1 & 96 & 1689 & 215 & 2489 & 0.809 & 0.773 & 0.737 & 0.000 & 0.146 & 0.284 & 0.634 & 0.814 & 12.053 & 0.461 & 0.244 & 0.216 & 0.787 \\
\texttt{one off   threshold} & 1 & 1 & 0 & 1 & 1 & 1 & 1 & 97 & 1694 & 213 & 2503 & 0.836 & 0.829 & 0.747 & 0.000 & 0.137 & 0.191 & 0.726 & 0.859 & 29.972 & 0.384 & 0.165 & 0.219 & 0.822 \\
\texttt{two off   threshold   structural} & 1 & 1 & 0 & 1 & 1 & 0 & 1 & 97 & 1694 & 213 & 2503 & 0.836 & 0.829 & 0.747 & 0.000 & 0.137 & 0.191 & 0.726 & 0.859 & 29.972 & 0.384 & 0.165 & 0.219 & 0.822 \\
\texttt{two off   threshold   decorrelation} & 1 & 1 & 0 & 1 & 1 & 1 & 0 & 96 & 1699 & 213 & 2503 & 0.836 & 0.829 & 0.747 & 0.000 & 0.139 & 0.190 & 0.734 & 0.865 & 30.002 & 0.388 & 0.166 & 0.222 & 0.821 \\
\texttt{two on   gain   homeostasis} & 0 & 1 & 0 & 0 & 1 & 0 & 0 & 96 & 1697 & 214 & 2513 & 0.810 & 0.774 & 0.738 & 0.000 & 0.147 & 0.283 & 0.642 & 0.820 & 12.069 & 0.462 & 0.244 & 0.218 & 0.787 \\
\texttt{two off   gain   inhibition} & 1 & 0 & 1 & 0 & 1 & 1 & 1 & 96 & 1712 & 213 & 2529 & 0.883 & 0.866 & 0.770 & 0.000 & 0.149 & 0.324 & 0.833 & 0.915 & 14.357 & 0.281 & 0.057 & 0.224 & 0.842 \\
\texttt{one off   input gate} & 0 & 1 & 1 & 1 & 1 & 1 & 1 & 97 & 1710 & 215 & 2537 & 0.802 & 0.775 & 0.729 & 0.000 & 0.144 & 0.290 & 0.615 & 0.801 & 12.011 & 0.440 & 0.236 & 0.204 & 0.789 \\
\texttt{two off   input gate   structural} & 0 & 1 & 1 & 1 & 1 & 0 & 1 & 97 & 1710 & 215 & 2537 & 0.802 & 0.775 & 0.729 & 0.000 & 0.144 & 0.290 & 0.615 & 0.801 & 12.011 & 0.440 & 0.236 & 0.204 & 0.789 \\
\texttt{two on   input gate   homeostasis} & 1 & 0 & 0 & 0 & 1 & 0 & 0 & 96 & 1709 & 215 & 2538 & 0.883 & 0.866 & 0.770 & 0.000 & 0.151 & 0.325 & 0.841 & 0.920 & 14.171 & 0.282 & 0.055 & 0.227 & 0.842 \\
\texttt{two off   input gate   threshold} & 0 & 1 & 0 & 1 & 1 & 1 & 1 & 97 & 1712 & 215 & 2540 & 0.804 & 0.776 & 0.729 & 0.000 & 0.143 & 0.290 & 0.616 & 0.802 & 12.023 & 0.436 & 0.233 & 0.202 & 0.791 \\
\texttt{two off   input gate   decorrelation} & 0 & 1 & 1 & 1 & 1 & 1 & 0 & 96 & 1714 & 215 & 2542 & 0.803 & 0.775 & 0.729 & 0.000 & 0.145 & 0.290 & 0.622 & 0.807 & 12.015 & 0.442 & 0.237 & 0.206 & 0.789 \\
\texttt{two off   gain   decorrelation} & 1 & 0 & 1 & 1 & 1 & 1 & 0 & 97 & 1719 & 215 & 2545 & 0.870 & 0.862 & 0.759 & 0.000 & 0.150 & 0.352 & 0.813 & 0.904 & 12.703 & 0.291 & 0.056 & 0.235 & 0.843 \\
\texttt{one off   gain} & 1 & 0 & 1 & 1 & 1 & 1 & 1 & 96 & 1725 & 215 & 2552 & 0.870 & 0.861 & 0.759 & 0.000 & 0.148 & 0.353 & 0.806 & 0.899 & 12.659 & 0.288 & 0.056 & 0.232 & 0.843 \\
\texttt{two off   gain   structural} & 1 & 0 & 1 & 1 & 1 & 0 & 1 & 96 & 1725 & 215 & 2552 & 0.870 & 0.861 & 0.759 & 0.000 & 0.148 & 0.353 & 0.806 & 0.899 & 12.659 & 0.288 & 0.056 & 0.232 & 0.843 \\
\texttt{two on   threshold   homeostasis} & 0 & 0 & 1 & 0 & 1 & 0 & 0 & 97 & 1722 & 213 & 2555 & 0.872 & 0.823 & 0.761 & 0.000 & 0.147 & 0.445 & 0.699 & 0.850 & 4.009 & 0.271 & 0.068 & 0.204 & 0.836 \\
\texttt{two off   gain   threshold} & 1 & 0 & 0 & 1 & 1 & 1 & 1 & 97 & 1726 & 214 & 2560 & 0.869 & 0.861 & 0.758 & 0.000 & 0.148 & 0.355 & 0.806 & 0.899 & 12.403 & 0.290 & 0.056 & 0.234 & 0.843 \\
\texttt{two on   homeostasis   decorrelation} & 0 & 0 & 0 & 0 & 1 & 0 & 1 & 96 & 1721 & 214 & 2570 & 0.872 & 0.823 & 0.761 & 0.000 & 0.145 & 0.447 & 0.694 & 0.846 & 3.924 & 0.269 & 0.066 & 0.203 & 0.837 \\
\texttt{one on   homeostasis} & 0 & 0 & 0 & 0 & 1 & 0 & 0 & 96 & 1723 & 214 & 2571 & 0.872 & 0.823 & 0.761 & 0.000 & 0.147 & 0.446 & 0.699 & 0.850 & 3.938 & 0.270 & 0.066 & 0.204 & 0.836 \\
\texttt{two on   homeostasis   structural} & 0 & 0 & 0 & 0 & 1 & 1 & 0 & 96 & 1723 & 214 & 2571 & 0.872 & 0.823 & 0.761 & 0.000 & 0.147 & 0.446 & 0.699 & 0.850 & 3.938 & 0.270 & 0.066 & 0.204 & 0.836 \\
\texttt{two off   input gate   gain} & 0 & 0 & 1 & 1 & 1 & 1 & 1 & 96 & 1728 & 214 & 2586 & 0.856 & 0.816 & 0.748 & 0.000 & 0.143 & 0.472 & 0.668 & 0.830 & 3.651 & 0.276 & 0.068 & 0.207 & 0.839 \\
\texttt{two on   inhibition   homeostasis} & 0 & 0 & 0 & 1 & 1 & 0 & 0 & 97 & 1733 & 215 & 2594 & 0.857 & 0.816 & 0.749 & 0.000 & 0.144 & 0.473 & 0.673 & 0.834 & 3.599 & 0.277 & 0.068 & 0.209 & 0.839 \\
\texttt{all off} & 0 & 0 & 0 & 0 & 0 & 0 & 0 & 862 & 0 & 994 & 0 & 0.181 & 0.374 & 0.284 & 0.000 & 0.495 & 0.453 & 0.105 & 0.431 & 2.012 & 0.868 & 0.828 & 0.039 & 0.195 \\
\texttt{two on   gain   decorrelation} & 0 & 1 & 0 & 0 & 0 & 0 & 1 & 921 & 0 & 998 & 0 & 0.131 & 0.373 & 0.239 & 0.000 & 0.445 & 0.462 & 0.082 & 0.404 & 2.034 & 0.890 & 0.858 & 0.033 & 0.138 \\
\texttt{one on   gain} & 0 & 1 & 0 & 0 & 0 & 0 & 0 & 921 & 0 & 998 & 0 & 0.184 & 0.387 & 0.282 & 0.000 & 0.488 & 0.446 & 0.104 & 0.431 & 2.044 & 0.864 & 0.825 & 0.039 & 0.198 \\
\texttt{two on   gain   structural} & 0 & 1 & 0 & 0 & 0 & 1 & 0 & 921 & 0 & 998 & 0 & 0.184 & 0.387 & 0.282 & 0.000 & 0.488 & 0.446 & 0.104 & 0.431 & 2.044 & 0.864 & 0.825 & 0.039 & 0.198 \\
\texttt{two on   gain   threshold} & 0 & 1 & 1 & 0 & 0 & 0 & 0 & 921 & 0 & 998 & 0 & 0.183 & 0.387 & 0.282 & 0.000 & 0.488 & 0.446 & 0.104 & 0.431 & 2.044 & 0.864 & 0.825 & 0.039 & 0.198 \\
\texttt{one on   decorrelation} & 0 & 0 & 0 & 0 & 0 & 0 & 1 & 922 & 0 & 999 & 0 & 0.131 & 0.364 & 0.240 & 0.000 & 0.451 & 0.467 & 0.083 & 0.405 & 2.051 & 0.891 & 0.858 & 0.033 & 0.138 \\
\texttt{two on   structural   decorrelation} & 0 & 0 & 0 & 0 & 0 & 1 & 1 & 922 & 0 & 999 & 0 & 0.131 & 0.364 & 0.240 & 0.000 & 0.451 & 0.467 & 0.083 & 0.405 & 2.051 & 0.891 & 0.858 & 0.033 & 0.138 \\
\texttt{one on   structural} & 0 & 0 & 0 & 0 & 0 & 1 & 0 & 922 & 0 & 999 & 0 & 0.181 & 0.377 & 0.282 & 0.000 & 0.493 & 0.451 & 0.105 & 0.432 & 2.061 & 0.866 & 0.826 & 0.039 & 0.195 \\
\texttt{two on   threshold   decorrelation} & 0 & 0 & 1 & 0 & 0 & 0 & 1 & 922 & 0 & 999 & 0 & 0.130 & 0.363 & 0.240 & 0.000 & 0.451 & 0.467 & 0.083 & 0.406 & 2.050 & 0.892 & 0.858 & 0.033 & 0.137 \\
\texttt{one on   threshold} & 0 & 0 & 1 & 0 & 0 & 0 & 0 & 922 & 0 & 999 & 0 & 0.181 & 0.377 & 0.282 & 0.000 & 0.493 & 0.451 & 0.105 & 0.432 & 2.061 & 0.867 & 0.827 & 0.039 & 0.194 \\
\texttt{two on   threshold   structural} & 0 & 0 & 1 & 0 & 0 & 1 & 0 & 922 & 0 & 999 & 0 & 0.181 & 0.377 & 0.282 & 0.000 & 0.493 & 0.451 & 0.105 & 0.432 & 2.061 & 0.867 & 0.827 & 0.039 & 0.194 \\
\texttt{two off   inhibition   homeostasis} & 1 & 1 & 1 & 0 & 0 & 1 & 1 & 923 & 0 & 1001 & 0 & 0.134 & 0.375 & 0.242 & 0.001 & 0.443 & 0.457 & 0.088 & 0.410 & 2.092 & 0.889 & 0.851 & 0.038 & 0.140 \\
\texttt{two on   input gate   gain} & 1 & 1 & 0 & 0 & 0 & 0 & 0 & 923 & 0 & 1001 & 0 & 0.187 & 0.387 & 0.284 & 0.000 & 0.485 & 0.441 & 0.111 & 0.438 & 2.103 & 0.863 & 0.818 & 0.044 & 0.200 \\
\texttt{two on   input gate   decorrelation} & 1 & 0 & 0 & 0 & 0 & 0 & 1 & 924 & 0 & 1002 & 0 & 0.134 & 0.365 & 0.243 & 0.001 & 0.448 & 0.461 & 0.089 & 0.412 & 2.116 & 0.889 & 0.850 & 0.039 & 0.141 \\
\texttt{two on   input gate   threshold} & 1 & 0 & 1 & 0 & 0 & 0 & 0 & 924 & 0 & 1002 & 0 & 0.185 & 0.377 & 0.284 & 0.000 & 0.490 & 0.445 & 0.112 & 0.439 & 2.127 & 0.863 & 0.818 & 0.045 & 0.199 \\
\texttt{one on   input gate} & 1 & 0 & 0 & 0 & 0 & 0 & 0 & 924 & 0 & 1002 & 0 & 0.184 & 0.377 & 0.284 & 0.000 & 0.490 & 0.445 & 0.112 & 0.439 & 2.127 & 0.865 & 0.820 & 0.045 & 0.197 \\
\texttt{two on   input gate   structural} & 1 & 0 & 0 & 0 & 0 & 1 & 0 & 924 & 0 & 1002 & 0 & 0.184 & 0.377 & 0.284 & 0.000 & 0.490 & 0.445 & 0.112 & 0.439 & 2.127 & 0.865 & 0.820 & 0.045 & 0.197 \\
\texttt{two off   input gate   homeostasis} & 0 & 1 & 1 & 1 & 0 & 1 & 1 & 930 & 0 & 1004 & 0 & 0.193 & 0.391 & 0.297 & 0.007 & 0.430 & 0.448 & 0.110 & 0.426 & 2.048 & 0.888 & 0.853 & 0.035 & 0.200 \\
\texttt{two on   gain   inhibition} & 0 & 1 & 0 & 1 & 0 & 0 & 0 & 929 & 0 & 1005 & 0 & 0.224 & 0.399 & 0.321 & 0.009 & 0.471 & 0.437 & 0.128 & 0.446 & 2.053 & 0.870 & 0.831 & 0.039 & 0.240 \\
\texttt{two on   inhibition   decorrelation} & 0 & 0 & 0 & 1 & 0 & 0 & 1 & 931 & 0 & 1006 & 0 & 0.192 & 0.381 & 0.297 & 0.006 & 0.435 & 0.453 & 0.110 & 0.427 & 2.066 & 0.887 & 0.852 & 0.036 & 0.200 \\
\texttt{two on   threshold   inhibition} & 0 & 0 & 1 & 1 & 0 & 0 & 0 & 930 & 0 & 1006 & 0 & 0.222 & 0.388 & 0.320 & 0.008 & 0.474 & 0.442 & 0.128 & 0.446 & 2.070 & 0.871 & 0.831 & 0.040 & 0.237 \\
\texttt{one on   inhibition} & 0 & 0 & 0 & 1 & 0 & 0 & 0 & 931 & 0 & 1006 & 0 & 0.221 & 0.388 & 0.320 & 0.008 & 0.474 & 0.442 & 0.128 & 0.447 & 2.070 & 0.871 & 0.832 & 0.040 & 0.235 \\
\texttt{two on   inhibition   structural} & 0 & 0 & 0 & 1 & 0 & 1 & 0 & 931 & 0 & 1006 & 0 & 0.221 & 0.388 & 0.320 & 0.008 & 0.474 & 0.442 & 0.128 & 0.447 & 2.070 & 0.871 & 0.832 & 0.040 & 0.235 \\
\texttt{two off   homeostasis   decorrelation} & 1 & 1 & 1 & 1 & 0 & 1 & 0 & 932 & 0 & 1008 & 0 & 0.227 & 0.399 & 0.323 & 0.009 & 0.467 & 0.431 & 0.134 & 0.453 & 2.115 & 0.869 & 0.824 & 0.044 & 0.241 \\
\texttt{one off   homeostasis} & 1 & 1 & 1 & 1 & 0 & 1 & 1 & 933 & 0 & 1008 & 0 & 0.196 & 0.392 & 0.300 & 0.007 & 0.427 & 0.442 & 0.116 & 0.433 & 2.112 & 0.885 & 0.845 & 0.040 & 0.204 \\
\texttt{two off   homeostasis   structural} & 1 & 1 & 1 & 1 & 0 & 0 & 1 & 933 & 0 & 1008 & 0 & 0.196 & 0.392 & 0.300 & 0.007 & 0.427 & 0.442 & 0.116 & 0.433 & 2.112 & 0.885 & 0.845 & 0.040 & 0.204 \\
\texttt{two off   threshold   homeostasis} & 1 & 1 & 0 & 1 & 0 & 1 & 1 & 933 & 0 & 1008 & 0 & 0.196 & 0.392 & 0.300 & 0.007 & 0.427 & 0.442 & 0.116 & 0.433 & 2.112 & 0.886 & 0.846 & 0.041 & 0.203 \\
\texttt{two on   input gate   inhibition} & 1 & 0 & 0 & 1 & 0 & 0 & 0 & 933 & 0 & 1009 & 0 & 0.224 & 0.389 & 0.322 & 0.008 & 0.470 & 0.436 & 0.135 & 0.454 & 2.142 & 0.870 & 0.824 & 0.045 & 0.238 \\
\texttt{two off   gain   homeostasis} & 1 & 0 & 1 & 1 & 0 & 1 & 1 & 934 & 0 & 1009 & 0 & 0.194 & 0.382 & 0.300 & 0.006 & 0.432 & 0.447 & 0.117 & 0.434 & 2.137 & 0.887 & 0.846 & 0.041 & 0.201 \\

\end{longtable}
\endgroup

\begingroup
\tiny
\setlength{\tabcolsep}{1.2pt}
\renewcommand{\arraystretch}{1.05}
\setlength{\LTleft}{0pt plus 1fill}
\setlength{\LTright}{0pt plus 1fill}

\begin{longtable}{@{}>{\raggedright\arraybackslash}p{4.35cm}*{25}{c}@{}}
\caption{Experimental results for the XOR structural density ablation with structural plasticity enabled at \(\rho_m=0.35\).}
\label{tab:xor-density-results}\\

\toprule
\textbf{Experiment} & \rotatebox{90}{\textbf{IG}} & \rotatebox{90}{\textbf{GM}} & \rotatebox{90}{\textbf{TM}} & \rotatebox{90}{\textbf{LI}} & \rotatebox{90}{\textbf{HO}} & \rotatebox{90}{\textbf{SP}} & \rotatebox{90}{\textbf{AD}} & \rotatebox{90}{\textbf{Train 0.9}} & \rotatebox{90}{\textbf{Test 0.9}} & \rotatebox{90}{\textbf{Train 1}} & \rotatebox{90}{\textbf{Test 1}} & \rotatebox{90}{\textbf{Lifetime Sparsity }} & \rotatebox{90}{\textbf{Sample Hoyer }} & \rotatebox{90}{\textbf{Neuron Hoyer }} & \rotatebox{90}{\textbf{Dead Neuron }} & \rotatebox{90}{\textbf{Mean Abs. Corr. }} & \rotatebox{90}{\textbf{Effective Rank }} & \rotatebox{90}{\textbf{Specialization }} & \rotatebox{90}{\textbf{Purity }} & \rotatebox{90}{\textbf{Centroid Sep. }} & \rotatebox{90}{\textbf{Within Jaccard }} & \rotatebox{90}{\textbf{Between Jaccard }} & \rotatebox{90}{\textbf{Jaccard Gap }} & \rotatebox{90}{\textbf{Activation Sparsity}} & \rotatebox{90}{\textbf{Structural Sparsity}} \\
\midrule
\endfirsthead

\caption[]{Experimental results for the XOR structural density ablation at \(\rho_m=0.35\) (continued).}\\
\toprule
\textbf{Experiment} & \rotatebox{90}{\textbf{IG}} & \rotatebox{90}{\textbf{GM}} & \rotatebox{90}{\textbf{TM}} & \rotatebox{90}{\textbf{LI}} & \rotatebox{90}{\textbf{HO}} & \rotatebox{90}{\textbf{SP}} & \rotatebox{90}{\textbf{AD}} & \rotatebox{90}{\textbf{Train 0.9}} & \rotatebox{90}{\textbf{Test 0.9}} & \rotatebox{90}{\textbf{Train 1}} & \rotatebox{90}{\textbf{Test 1}} & \rotatebox{90}{\textbf{Lifetime Sparsity }} & \rotatebox{90}{\textbf{Sample Hoyer }} & \rotatebox{90}{\textbf{Neuron Hoyer }} & \rotatebox{90}{\textbf{Dead Neuron }} & \rotatebox{90}{\textbf{Mean Abs. Corr. }} & \rotatebox{90}{\textbf{Effective Rank }} & \rotatebox{90}{\textbf{Specialization }} & \rotatebox{90}{\textbf{Purity }} & \rotatebox{90}{\textbf{Centroid Sep. }} & \rotatebox{90}{\textbf{Within Jaccard }} & \rotatebox{90}{\textbf{Between Jaccard }} & \rotatebox{90}{\textbf{Jaccard Gap }} & \rotatebox{90}{\textbf{Activation Sparsity}} & \rotatebox{90}{\textbf{Structural Sparsity}} \\
\midrule
\endhead

\midrule
\multicolumn{26}{r}{Continued on next page}\\
\endfoot

\bottomrule
\endlastfoot

\texttt{two off   gain   decorrelation} & 1 & 0 & 1 & 1 & 1 & 1 & 0 & 160 & 1427 & 310 & 2353 & 0.840 & 0.708 & 0.744 & 0.000 & 0.104 & 0.723 & 0.435 & 0.632 & 0.517 & 0.170 & 0.068 & 0.103 & 0.864 & 0.65 \\
\texttt{two off   input gate   gain} & 0 & 0 & 1 & 1 & 1 & 1 & 1 & 158 & 1445 & 310 & 2610 & 0.840 & 0.709 & 0.744 & 0.000 & 0.102 & 0.724 & 0.418 & 0.623 & 0.510 & 0.167 & 0.069 & 0.098 & 0.864 & 0.65 \\
\texttt{two off   gain   inhibition} & 1 & 0 & 1 & 0 & 1 & 1 & 1 & 157 & 1418 & 314 & 2818 & 0.849 & 0.717 & 0.752 & 0.000 & 0.104 & 0.711 & 0.444 & 0.637 & 0.512 & 0.166 & 0.063 & 0.103 & 0.864 & 0.65 \\
\texttt{one off   gain} & 1 & 0 & 1 & 1 & 1 & 1 & 1 & 158 & 1431 & 307 & 2966 & 0.840 & 0.709 & 0.744 & 0.000 & 0.102 & 0.724 & 0.429 & 0.629 & 0.515 & 0.169 & 0.068 & 0.101 & 0.864 & 0.65 \\
\texttt{one off   threshold} & 1 & 1 & 0 & 1 & 1 & 1 & 1 & 159 & 1424 & 311 & 3012 & 0.829 & 0.847 & 0.735 & 0.000 & 0.106 & 0.466 & 0.429 & 0.628 & 14.762 & 0.202 & 0.070 & 0.132 & 0.856 & 0.65 \\
\texttt{two off   gain   threshold} & 1 & 0 & 0 & 1 & 1 & 1 & 1 & 158 & 1442 & 313 & 3029 & 0.841 & 0.709 & 0.745 & 0.000 & 0.102 & 0.724 & 0.430 & 0.629 & 0.514 & 0.169 & 0.068 & 0.101 & 0.864 & 0.65 \\
\texttt{one off   decorrelation} & 1 & 1 & 1 & 1 & 1 & 1 & 0 & 160 & 1431 & 314 & 3036 & 0.828 & 0.847 & 0.734 & 0.000 & 0.108 & 0.460 & 0.433 & 0.632 & 15.454 & 0.204 & 0.071 & 0.133 & 0.855 & 0.65 \\
\texttt{all on} & 1 & 1 & 1 & 1 & 1 & 1 & 1 & 158 & 1447 & 310 & 3081 & 0.829 & 0.847 & 0.735 & 0.000 & 0.106 & 0.463 & 0.429 & 0.629 & 15.058 & 0.202 & 0.070 & 0.132 & 0.856 & 0.65 \\
\texttt{two off   input gate   decorrelation} & 0 & 1 & 1 & 1 & 1 & 1 & 0 & 160 & 1423 & 310 & 3091 & 0.828 & 0.825 & 0.734 & 0.000 & 0.107 & 0.539 & 0.419 & 0.625 & 7.906 & 0.200 & 0.071 & 0.128 & 0.855 & 0.65 \\
\texttt{two off   threshold   decorrelation} & 1 & 1 & 0 & 1 & 1 & 1 & 0 & 159 & 1440 & 313 & 3096 & 0.828 & 0.846 & 0.733 & 0.000 & 0.108 & 0.463 & 0.433 & 0.631 & 15.082 & 0.203 & 0.070 & 0.133 & 0.855 & 0.65 \\
\texttt{one off   inhibition} & 1 & 1 & 1 & 0 & 1 & 1 & 1 & 158 & 1420 & 313 & 3097 & 0.837 & 0.854 & 0.742 & 0.000 & 0.108 & 0.433 & 0.445 & 0.638 & 17.618 & 0.203 & 0.066 & 0.137 & 0.854 & 0.65 \\
\texttt{two off   input gate   inhibition} & 0 & 1 & 1 & 0 & 1 & 1 & 1 & 157 & 1442 & 312 & 3103 & 0.838 & 0.833 & 0.742 & 0.000 & 0.107 & 0.515 & 0.431 & 0.630 & 8.864 & 0.198 & 0.067 & 0.132 & 0.855 & 0.65 \\
\texttt{two off   threshold   inhibition} & 1 & 1 & 0 & 0 & 1 & 1 & 1 & 159 & 1425 & 314 & 3105 & 0.837 & 0.854 & 0.742 & 0.000 & 0.108 & 0.436 & 0.444 & 0.637 & 17.290 & 0.202 & 0.066 & 0.136 & 0.854 & 0.65 \\
\texttt{two on   homeostasis   structural} & 0 & 0 & 0 & 0 & 1 & 1 & 0 & 156 & 1432 & 314 & 3115 & 0.848 & 0.716 & 0.751 & 0.000 & 0.104 & 0.711 & 0.437 & 0.635 & 0.509 & 0.164 & 0.064 & 0.100 & 0.863 & 0.65 \\
\texttt{two off   inhibition   decorrelation} & 1 & 1 & 1 & 0 & 1 & 1 & 0 & 157 & 1426 & 311 & 3124 & 0.837 & 0.855 & 0.741 & 0.000 & 0.109 & 0.430 & 0.450 & 0.641 & 18.046 & 0.204 & 0.065 & 0.138 & 0.854 & 0.65 \\
\texttt{one off   input gate} & 0 & 1 & 1 & 1 & 1 & 1 & 1 & 159 & 1432 & 309 & 3137 & 0.829 & 0.826 & 0.735 & 0.000 & 0.105 & 0.540 & 0.415 & 0.622 & 7.813 & 0.198 & 0.071 & 0.127 & 0.856 & 0.65 \\
\texttt{two off   input gate   threshold} & 0 & 1 & 0 & 1 & 1 & 1 & 1 & 157 & 1435 & 304 & 3138 & 0.829 & 0.826 & 0.735 & 0.000 & 0.105 & 0.542 & 0.415 & 0.622 & 7.718 & 0.198 & 0.071 & 0.127 & 0.856 & 0.65 \\
\texttt{two off   homeostasis   decorrelation} & 1 & 1 & 1 & 1 & 0 & 1 & 0 & 1035 & 3740 & 1245 & 3992 & 0.315 & 0.501 & 0.388 & 0.121 & 0.415 & 0.179 & 0.184 & 0.426 & 25.931 & 0.892 & 0.816 & 0.076 & 0.331 & 0.65 \\
\texttt{one off   homeostasis} & 1 & 1 & 1 & 1 & 0 & 1 & 1 & 1036 & 3763 & 1245 & 4000 & 0.302 & 0.498 & 0.376 & 0.114 & 0.393 & 0.183 & 0.173 & 0.417 & 25.824 & 0.901 & 0.828 & 0.073 & 0.315 & 0.65 \\
\texttt{two off   threshold   homeostasis} & 1 & 1 & 0 & 1 & 0 & 1 & 1 & 1036 & 3763 & 1244 & 4000 & 0.301 & 0.498 & 0.376 & 0.114 & 0.392 & 0.183 & 0.174 & 0.418 & 25.827 & 0.901 & 0.828 & 0.073 & 0.314 & 0.65 \\
\texttt{two off   gain   homeostasis} & 1 & 0 & 1 & 1 & 0 & 1 & 1 & 1038 & 3793 & 1246 & 4052 & 0.301 & 0.486 & 0.376 & 0.113 & 0.396 & 0.193 & 0.173 & 0.418 & 23.719 & 0.900 & 0.827 & 0.072 & 0.314 & 0.65 \\
\texttt{two off   inhibition   homeostasis} & 1 & 1 & 1 & 0 & 0 & 1 & 1 & 1027 & 3975 & 1264 & 4122 & 0.249 & 0.473 & 0.325 & 0.051 & 0.363 & 0.198 & 0.171 & 0.430 & 23.834 & 0.901 & 0.835 & 0.066 & 0.261 & 0.65 \\
\texttt{two on   input gate   structural} & 1 & 0 & 0 & 0 & 0 & 1 & 0 & 1025 & 3968 & 1265 & 4168 & 0.276 & 0.468 & 0.350 & 0.035 & 0.375 & 0.201 & 0.202 & 0.462 & 22.044 & 0.889 & 0.819 & 0.069 & 0.291 & 0.65 \\
\texttt{two off   input gate   homeostasis} & 0 & 1 & 1 & 1 & 0 & 1 & 1 & 1036 & 4508 & 1242 & 4996 & 0.291 & 0.448 & 0.369 & 0.111 & 0.421 & 0.277 & 0.143 & 0.398 & 9.441 & 0.902 & 0.841 & 0.062 & 0.302 & 0.65 \\
\texttt{two on   inhibition   structural} & 0 & 0 & 0 & 1 & 0 & 1 & 0 & 1035 & 4554 & 1243 & 0 & 0.305 & 0.441 & 0.383 & 0.115 & 0.447 & 0.281 & 0.157 & 0.409 & 8.916 & 0.895 & 0.830 & 0.065 & 0.318 & 0.65 \\
\texttt{two on   gain   structural} & 0 & 1 & 0 & 0 & 0 & 1 & 0 & 1024 & 4433 & 1261 & 0 & 0.275 & 0.436 & 0.349 & 0.038 & 0.400 & 0.283 & 0.176 & 0.444 & 8.875 & 0.888 & 0.829 & 0.059 & 0.289 & 0.65 \\
\texttt{two on   threshold   structural} & 0 & 0 & 1 & 0 & 0 & 1 & 0 & 1025 & 4468 & 1262 & 0 & 0.272 & 0.425 & 0.350 & 0.038 & 0.404 & 0.293 & 0.177 & 0.444 & 8.375 & 0.893 & 0.831 & 0.062 & 0.284 & 0.65 \\
\texttt{two on   structural   decorrelation} & 0 & 0 & 0 & 0 & 0 & 1 & 1 & 1028 & 4472 & 1262 & 0 & 0.245 & 0.417 & 0.324 & 0.041 & 0.388 & 0.301 & 0.157 & 0.424 & 8.353 & 0.902 & 0.845 & 0.057 & 0.256 & 0.65 \\
\texttt{one on   structural} & 0 & 0 & 0 & 0 & 0 & 1 & 0 & 1025 & 4464 & 1263 & 0 & 0.273 & 0.425 & 0.349 & 0.038 & 0.404 & 0.293 & 0.177 & 0.444 & 8.374 & 0.891 & 0.830 & 0.061 & 0.286 & 0.65 \\

\end{longtable}
\endgroup

\end{landscape}

Because the noisy XOR inputs are 40,000-dimensional while the class-relevant structure is concentrated in only two signal directions, the density of the structural mask can materially change the balance between suppressing incidental high-dimensional variation and preserving the connectivity needed to represent the four-cluster geometry. The main XOR ablation in Table~\ref{tab:xor-results} used \(\rho_m=1\), which leaves all connections active. To test whether the conclusions about structural plasticity depend on this dense setting, we repeated every XOR configuration using the structural mask with \(\rho_m=0.35\), the same density used for sparse parity, while leaving all other experimental settings unchanged. The additional results are reported in Table~\ref{tab:xor-density-results}; their comparison with the original XOR ablations is discussed separately below.

To complement the aggregate results in the tables, the following learning curves in Figures 2--5 illustrate the grokking dynamics for four representative sparse parity configurations. The all-off and all-on cases show the two extremes of the ablation study, while the other two plots show the best configurations, according to ``Test 1'', with only two mechanisms enabled and with only one mechanism enabled. These are homeostasis with structural plasticity, and structural plasticity alone, respectively. Since the quantitative differences are discussed below, the purpose of these figures is mainly to visualize the temporal separation between fitting the training set and generalizing to the test set. The solid lines show mean accuracy over 20 independent runs, and the shaded regions indicate $\pm 1$ standard deviation.

\begin{figure}[t]
    \centering
    \includegraphics[width=\linewidth]{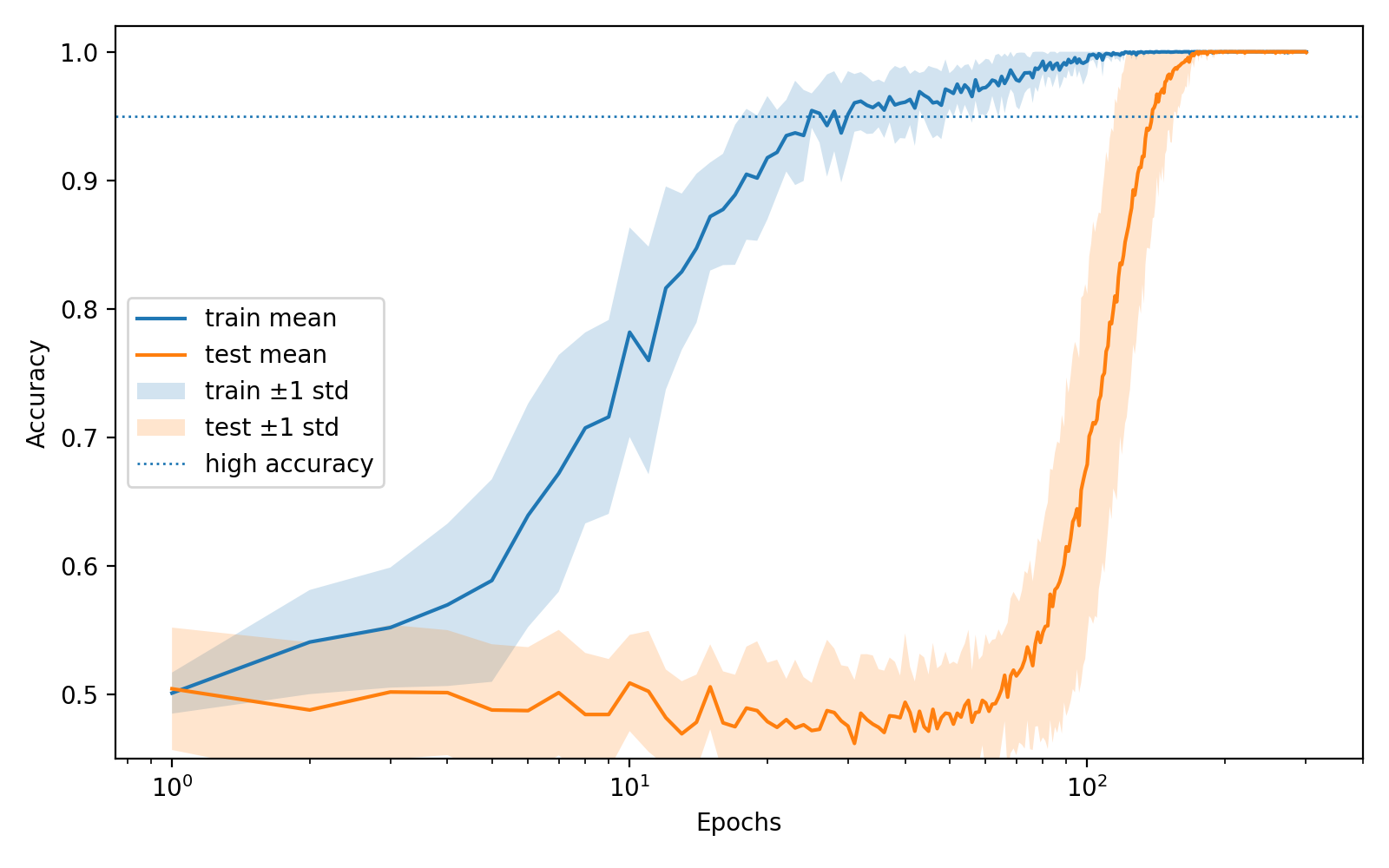}
    \caption{Sparse parity with all bio-inspired mechanisms disabled. Training accuracy improves well before test accuracy, which produces a pronounced delayed generalization phase.}
    \label{fig:parity_all_off}
\end{figure}

\begin{figure}[t]
    \centering
    \includegraphics[width=\linewidth]{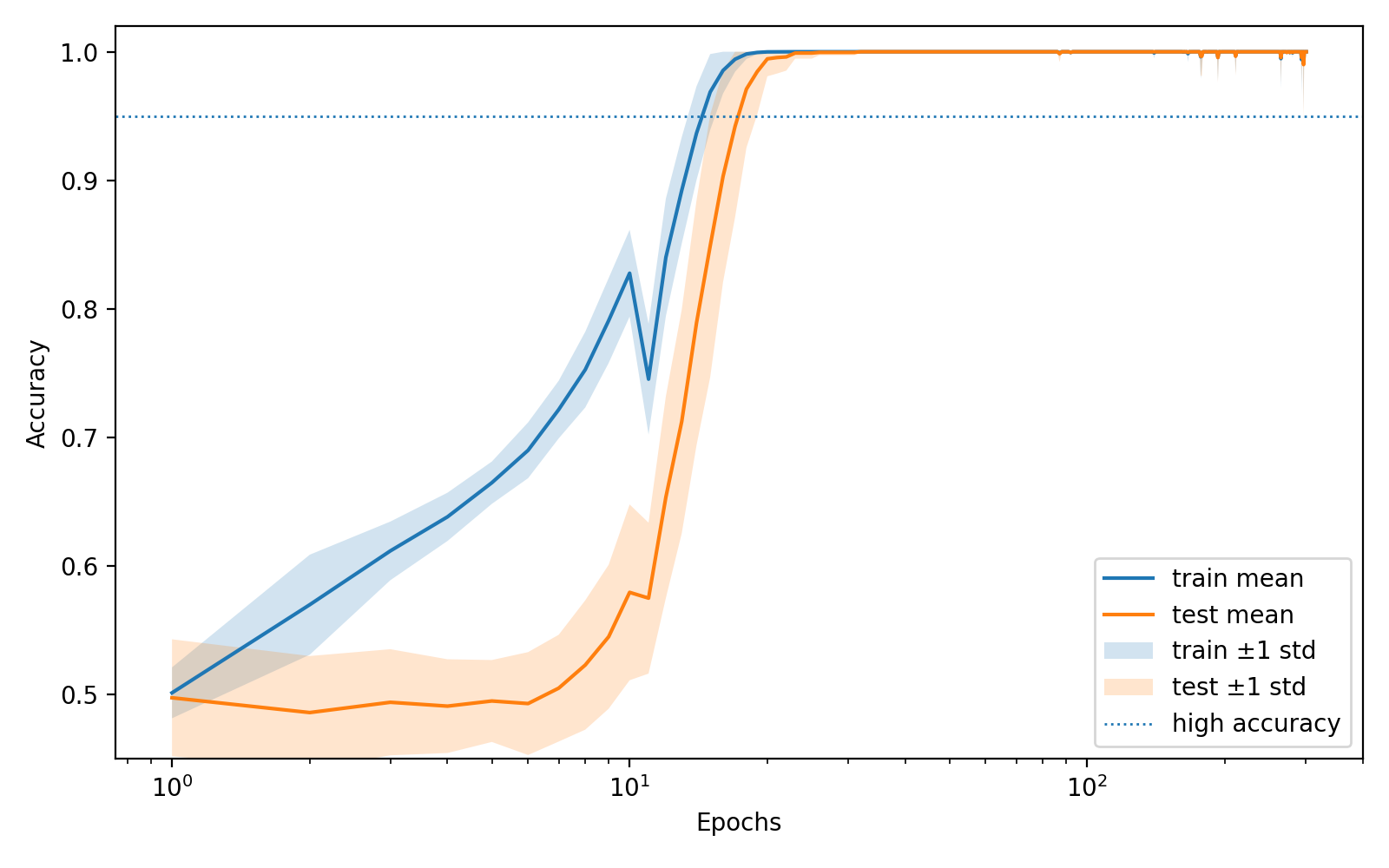}
    \caption{Sparse parity with all bio-inspired mechanisms enabled. Training and test accuracy rise in close succession, which results in substantially faster grokking than in the all-off case.}
    \label{fig:parity_all_on}
\end{figure}

\begin{figure}[t]
    \centering
    \includegraphics[width=\linewidth]{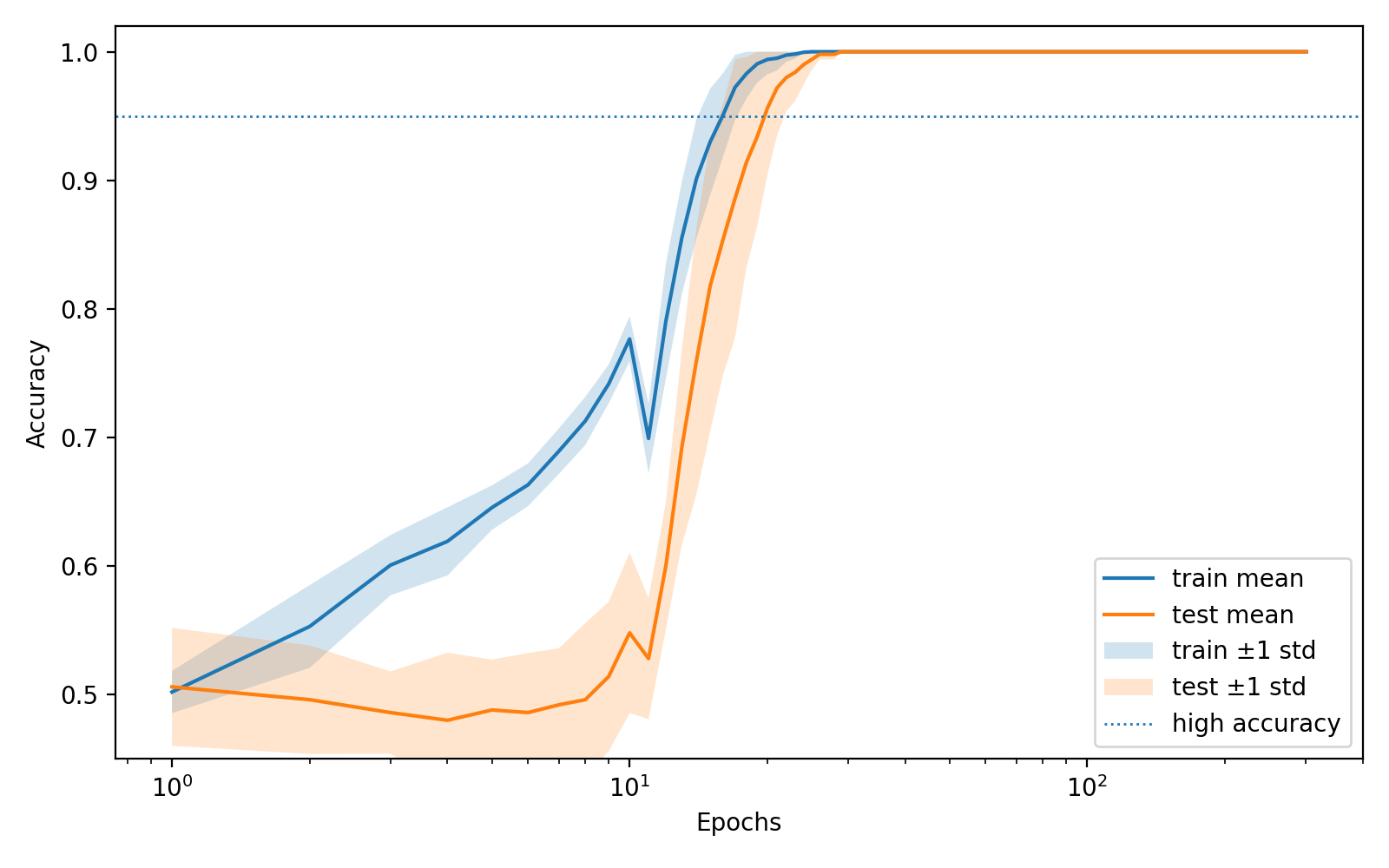}
    \caption{Sparse parity with only homeostasis and structural plasticity enabled. This best two-mechanism configuration shows rapid generalization and closely reproduces the fast grokking behavior of the fully enabled model.}
    \label{fig:parity_two_on_homeostasis_structural}
\end{figure}

\begin{figure}[t]
    \centering
    \includegraphics[width=\linewidth]{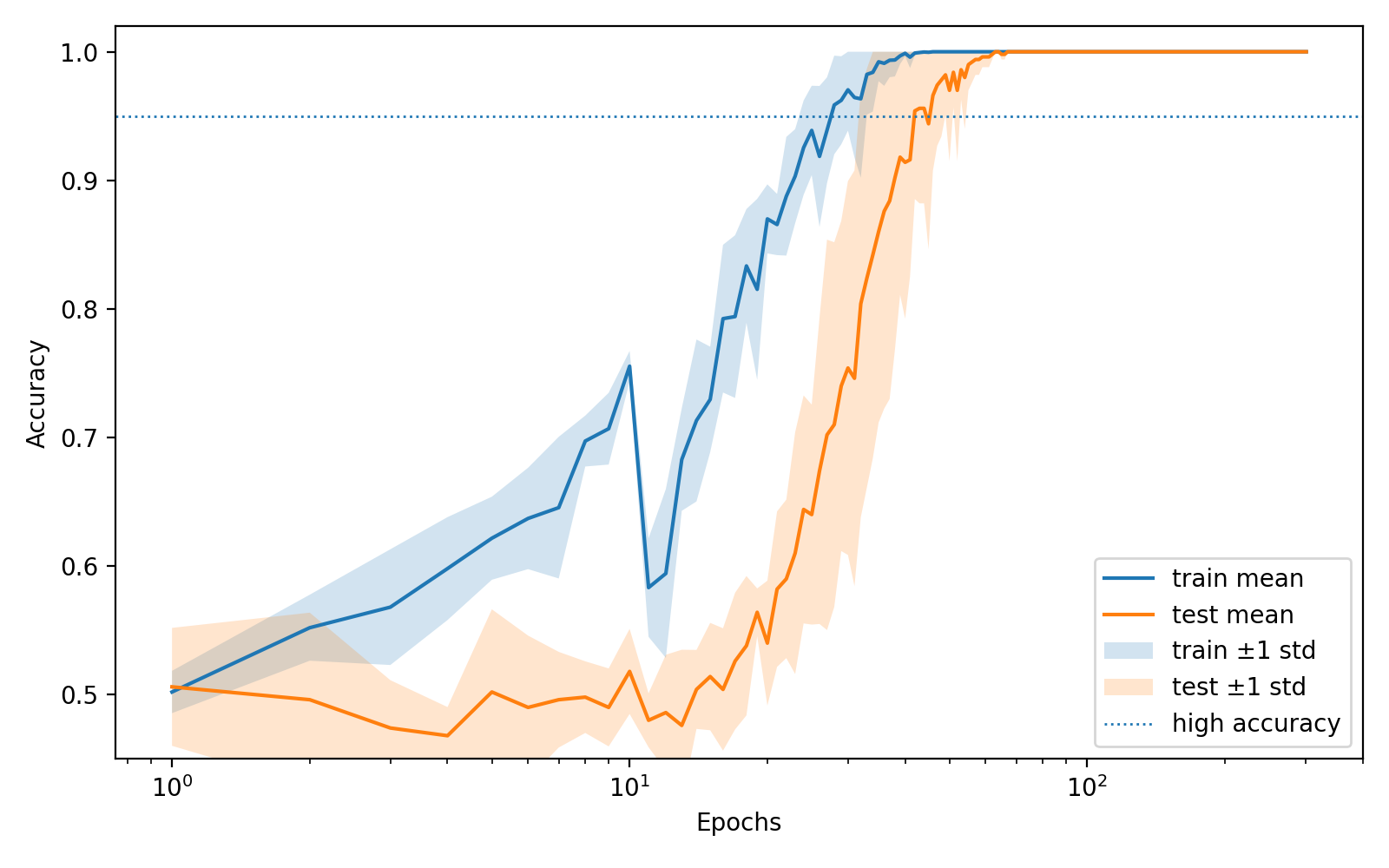}
    \caption{Sparse parity with only structural plasticity enabled. Generalization remains delayed relative to training, but the transition occurs much earlier than in the all-off case.}
    \label{fig:parity_one_on_structural}
\end{figure}

For noisy XOR, the selected learning curves emphasize the contrast between memorization without generalization and successful delayed grokking. In addition to the all-off and all-on configurations, we show the two best two-mechanism configurations according to ``Test 1'': gain modulation with homeostasis, and input gating with homeostasis. No single mechanism configuration approaches the all-on result closely enough to provide an equally informative comparison, so single mechanism cases are omitted here. The plotting conventions are the same as for the sparse parity figures.

\begin{figure}[t]
    \centering
    \includegraphics[width=\linewidth]{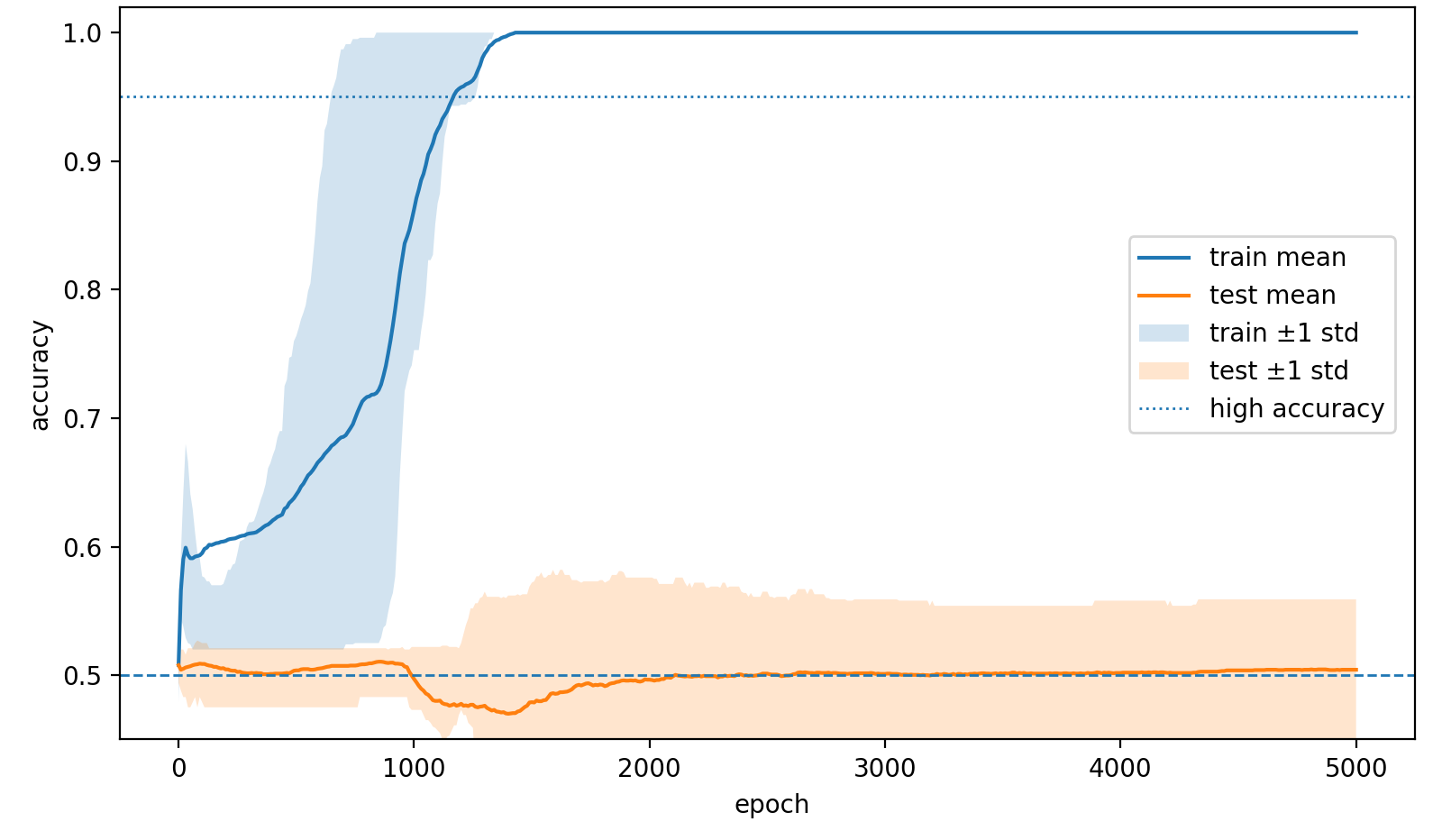}
    \caption{Noisy XOR with all bio-inspired mechanisms disabled. Training accuracy eventually reaches 1, but test accuracy remains near chance throughout the experimental horizon, which indicates memorization without generalization.}
    \label{fig:xor_all_off}
\end{figure}

\begin{figure}[t]
    \centering
    \includegraphics[width=\linewidth]{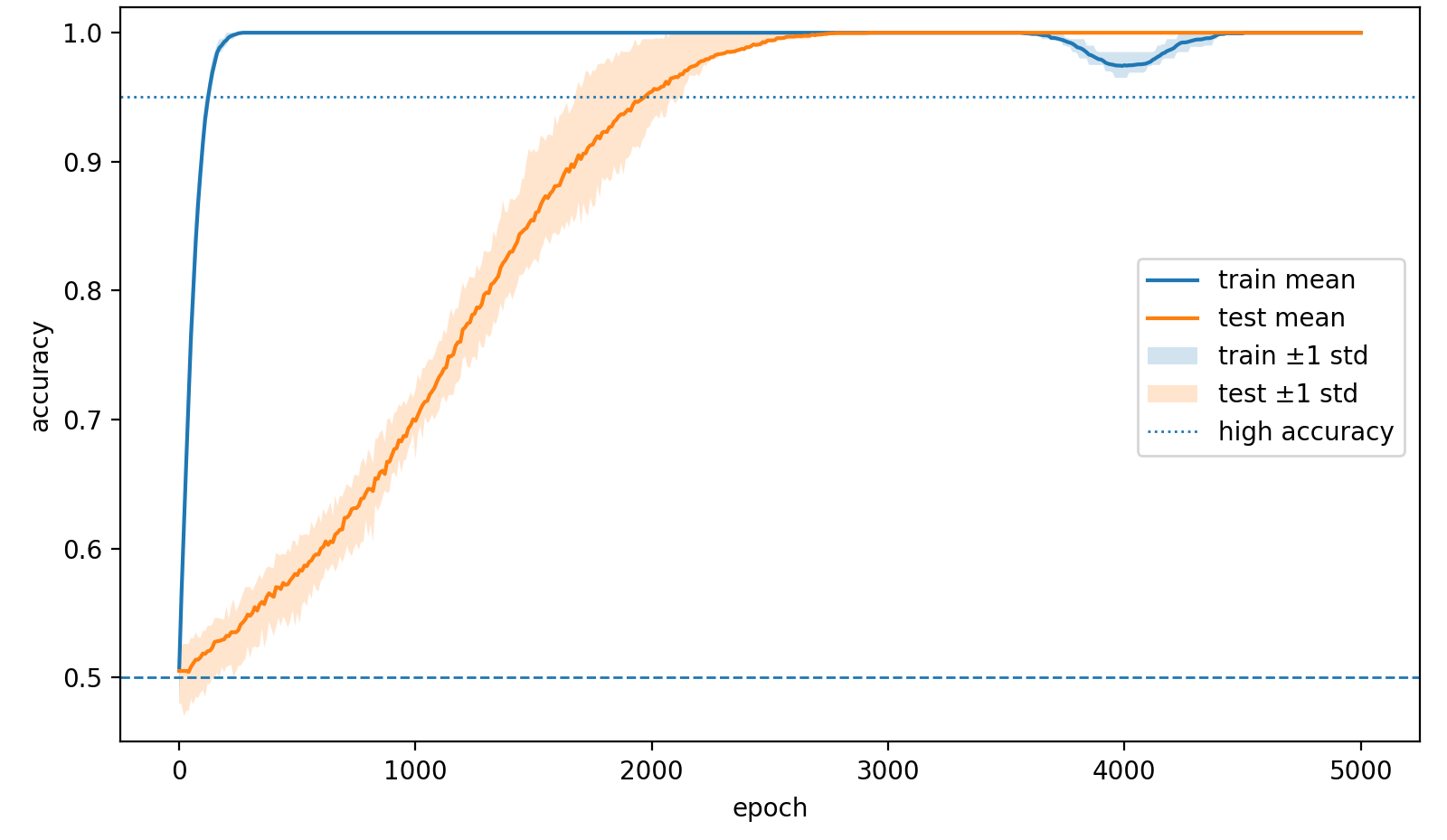}
    \caption{Noisy XOR with all bio-inspired mechanisms enabled. Training accuracy reaches its maximum early, whereas test accuracy improves progressively over a much longer interval before reaching near-perfect performance. The separation between these two timescales illustrates delayed generalization.}
    \label{fig:xor_all_on}
\end{figure}

\begin{figure}[t]
    \centering
    \includegraphics[width=\linewidth]{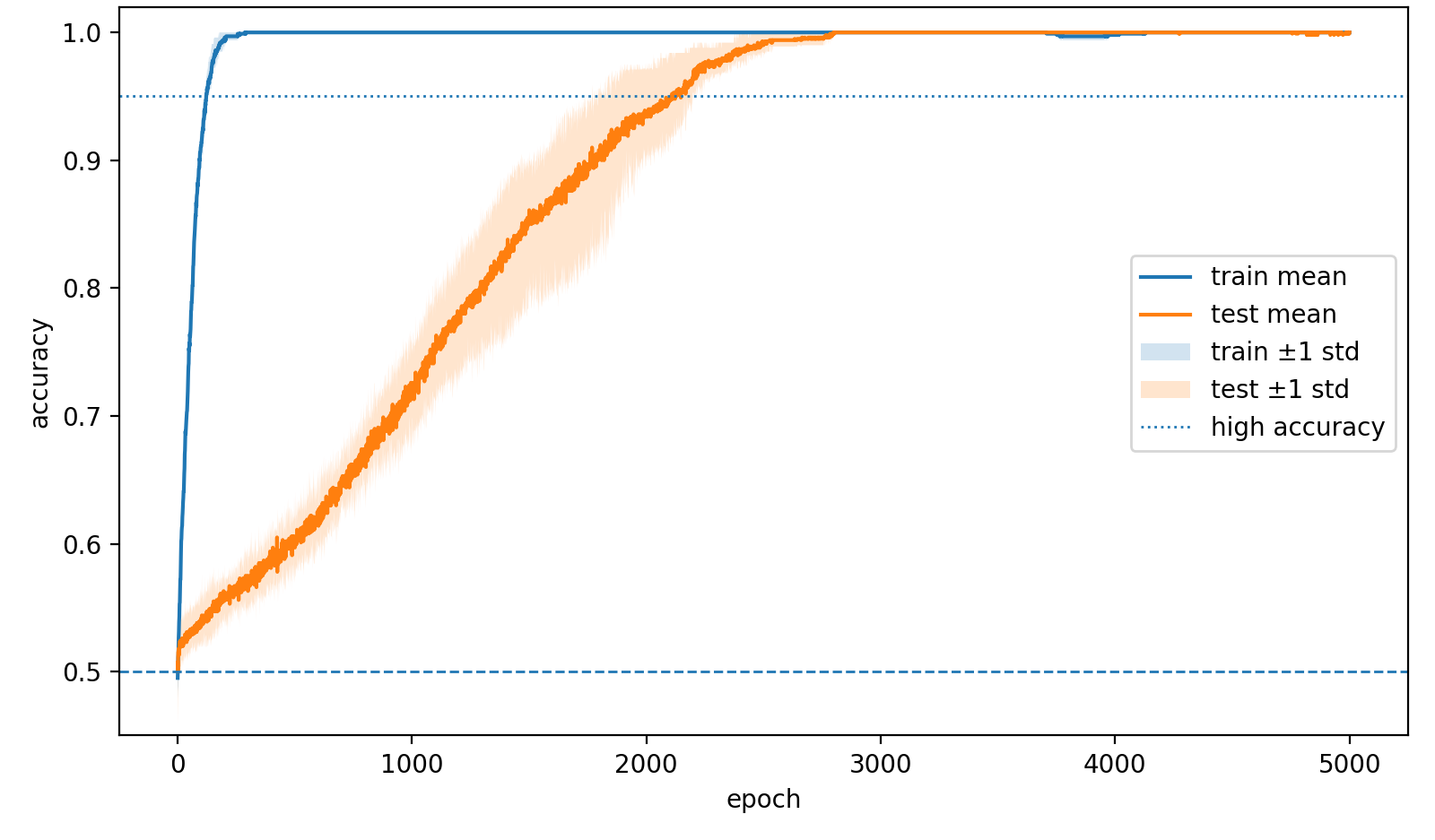}
    \caption{Noisy XOR with only gain modulation and homeostasis enabled. This configuration is the best two-mechanism case according to ``Test 1''.}
    \label{fig:xor_two_on_gain_homeostasis}
\end{figure}

\begin{figure}[t]
    \centering
    \includegraphics[width=\linewidth]{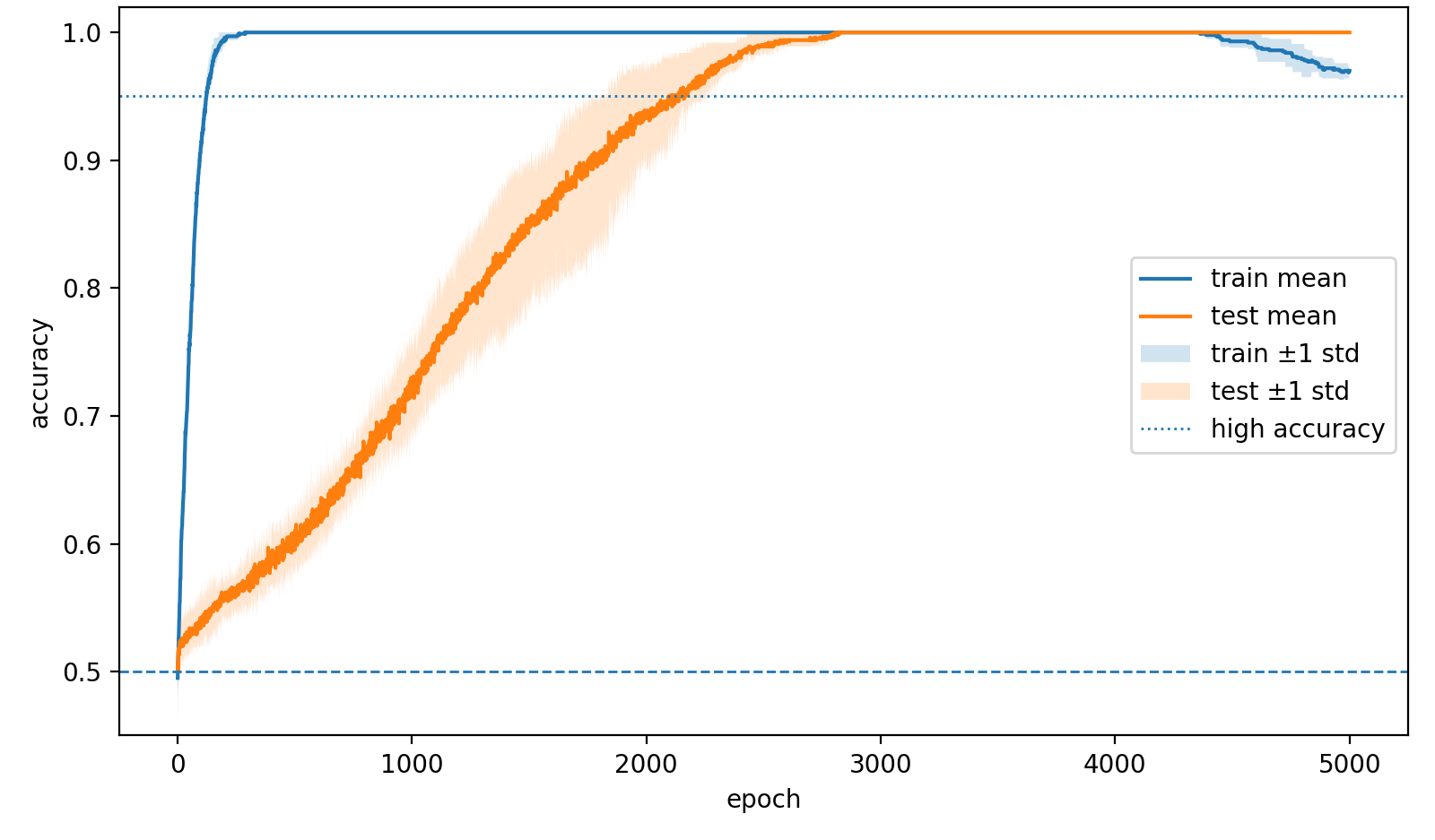}
    \caption{Noisy XOR with only input gating and homeostasis enabled. This configuration is the second-best two-mechanism case according to ``Test 1''.}
    \label{fig:xor_two_on_input_gate_homeostasis}
\end{figure}

The ablation results show that the seven biologically inspired mechanisms do not contribute equally to grokking. The clearest pattern is that only a small subset of mechanisms explains most of the improvement. The dominant mechanisms are homeostasis and structural plasticity, although their roles differ between the two tasks.

The sparse parity task is governed mainly by the formation of a compact sparse circuit. In this setting, structural plasticity and homeostasis are the two decisive mechanisms. The XOR task shows a different but related pattern. The XOR task shows a different but related pattern. Homeostasis provides the most consistent route to generalization, while the structural-density ablation shows that sparse effective connectivity can itself produce a major generalization benefit. For both tasks, the strongest effects therefore come from mechanisms that regulate neuron use and effective connectivity.

\subsection{Discussion: Sparse Parity}

The sparse parity results strongly support the view that grokking is accelerated when the network forms a small effective subnetwork.

The all-off baseline reaches perfect training accuracy at epoch 70 and perfect test accuracy at epoch 137. The fully enabled model reaches perfect training accuracy at epoch 16 and perfect test accuracy at epoch 18. This is a large reduction in the grokking delay. However, the best configurations are not the fully enabled ones. The best parity result removes lateral inhibition and activation decorrelation and reaches perfect test accuracy at epoch 17. Other configurations that remove gain modulation, threshold modulation, lateral inhibition, or decorrelation also remain in the same fast-generalization regime.

This means that the parity result is not caused by the presence of all seven mechanisms. Instead, the top parity rows share two essential properties: homeostasis is enabled and structural plasticity is enabled. In the top fifteen parity configurations, both homeostasis and structural plasticity are present in every case. In the bottom fifteen configurations, both are absent. This is the clearest signal in the parity ablation.

Structural plasticity has the largest individual effect. Removing structural plasticity from the full model delays Test 1 from 18 to 53. Using structural plasticity alone reaches Test 1 = 43, which is much better than the all-off baseline. Structural plasticity therefore carries a substantial part of the parity improvement even without the other mechanisms.

Homeostasis is the second major mechanism. Removing homeostasis from the full model delays Test1 from 18 to 35. Using homeostasis alone reaches Test 1 = 69, which is also a clear improvement over the baseline. The strongest evidence comes from the two-mechanism configuration with homeostasis and structural plasticity. This configuration reaches Train 1 = 19 and Test 1 = 22, close to the full model and close to the best parity rows overall. Thus, the pair homeostasis plus structural plasticity is nearly sufficient for the parity effect.
The internal metrics support this interpretation. The best parity configurations have high activation sparsity, high Hoyer sparsity, low effective rank, and a small final active subnetwork. In the top fifteen parity runs, average activation sparsity is about 0.882, compared with about 0.584 in the bottom fifteen runs. Average sample Hoyer sparsity is about 0.996 in the top group and about 0.801 in the bottom group. Average neuron Hoyer sparsity rises from about 0.498 to about 0.857. Effective rank falls from about 0.089 to about 0.009.

The dead neuron fraction also changes substantially. In the best parity runs, many hidden neurons become inactive, with an average dead-neuron fraction of about 0.713. In the weakest runs, the value is essentially zero. This should not be read as a failure of representation. In sparse parity, a high dead-neuron fraction indicates that the network has compressed the computation into a small set of useful hidden neurons. The final active subnetwork size remains very small in the strongest configurations. The best row has a final active subnetwork size of about 7.4 neurons, and the homeostasis-plus-structural-plasticity configuration has about 7.2 active neurons.

The parity result therefore has a direct interpretation. Structural plasticity selects the effective connections that participate in the rule-like computation. Homeostasis regulates how frequently neurons are used during the formation of that circuit. Together, they push the model toward a compact solution that generalizes much earlier than the dense baseline.

The remaining mechanisms are secondary for sparse parity. Input gating has a modest positive effect in the full model, but it is not essential once homeostasis and structural plasticity are present. Removing input gating from the full model gives Test 1 = 20, still close to the best regime. Gain modulation, threshold modulation, lateral inhibition, and activation decorrelation are even less central in this task. Removing each of them individually does not harm parity convergence and can slightly improve it. These mechanisms may still change the details of the hidden representation, but they do not explain the main grokking improvement.

\subsection{Discussion: Noisy XOR}

The all-off baseline reaches perfect training accuracy at step 994 but never reaches perfect test accuracy within the run. The fully enabled model reaches perfect training accuracy at step 219 and perfect test accuracy at step 2310. This shows that the full model both learns the training set earlier and eventually reaches perfect test generalization, while the baseline remains stuck without test grokking.

The original XOR ablation identifies homeostasis as the dominant mechanism. At
\(\rho_m=1\), every configuration that reaches perfect test accuracy contains
homeostasis, and homeostasis alone reaches Test 1 = 2571. The two-mechanism
configurations show the same pattern: gain modulation plus homeostasis reaches
Test 1 = 2513, input gating plus homeostasis reaches Test 1 = 2538, and threshold
modulation plus homeostasis reaches Test 1 = 2555. Thus, under the dense structural
setting used in the main XOR ablation, homeostatic regulation is the mechanism most
consistently associated with successful generalization.

The structural-density ablation in Table~\ref{tab:xor-density-results} qualifies this
conclusion. When structural plasticity is active with \(\rho_m=0.35\), configurations
without homeostasis can also generalize. Among the twelve repeated configurations
without homeostasis, all reach Test 0.9 and seven reach perfect test accuracy, whereas
none reaches either threshold in the corresponding \(\rho_m=1\) runs. Structural
sparsification can therefore provide an alternative route to XOR generalization when
homeostatic regulation is absent.

The interaction changes again when homeostasis is present. In these configurations,
\(\rho_m=0.35\) generally produces an earlier transition to high test accuracy, but
does not improve final convergence: Test 0.9 is reached earlier in all seventeen
matched configurations, while Test 1 is reached later in sixteen of the seventeen.
The effect of structural plasticity in XOR is therefore conditional on the regulatory
context. Homeostasis remains the most reliable mechanism across structural densities,
while stronger structural sparsification is particularly beneficial when homeostatic
regulation is unavailable.

The internal metrics clarify how the two regulatory mechanisms affect the hidden
representation. In the original dense structural setting, the failed XOR runs have weak
latent-regime organization. The all-off baseline has activation sparsity about 0.195,
specialization about 0.105, purity about 0.431, centroid separation about 2.01, and a
Jaccard gap about 0.039. By comparison, the fully enabled model has activation sparsity
about 0.818, specialization about 0.720, purity about 0.858, centroid separation about
28.78, and a Jaccard gap about 0.222. The original ablation therefore associates successful
generalization with a substantially more selective and group-specific hidden representation.

The structural-density ablation shows that this organization is not unique. When
homeostasis is absent, reducing the structural density to \(\rho_m=0.35\) consistently
makes the representation sparser and less correlated, while increasing specialization,
centroid separation, and the separation between within-group and between-group active
supports. These changes accompany the emergence of test generalization that is absent
from the corresponding dense runs. Structural sparsification can therefore organize the
XOR representation sufficiently for generalization even without homeostatic regulation.

When homeostasis is already active, however, the same structural constraint produces a
different effect. The \(\rho_m=0.35\) runs again have sparser and less correlated activity,
but specialization, purity, centroid separation, and the Jaccard gap decrease across the
matched configurations, while effective rank increases. The models nevertheless
generalize, although they usually reach perfect test accuracy later. Thus, successful XOR
generalization is not associated with a single representational signature. Homeostasis and
structural sparsification appear to impose different, partly overlapping forms of internal
organization, and their interaction determines which representation emerges. 

The remaining mechanisms have more limited effects on XOR generalization. Gain modulation
has the strongest complementary effect in the original two-mechanism ablation: gain
modulation plus homeostasis reaches Test 1 = 2513, compared with Test 1 = 2571 for
homeostasis alone. Input gating plus homeostasis also performs well, and its internal
metrics show strong specialization and purity. Removing gain modulation, input gating,
threshold modulation, lateral inhibition, or activation decorrelation individually from the
fully enabled model does not prevent generalization, indicating that none of these
mechanisms is individually required for the XOR task.

Structural plasticity should be interpreted separately from these secondary mechanisms.
With \(\rho_m=1\), removing it from the fully enabled model has only a modest effect,
consistent with the fact that this setting leaves the effective connectivity dense. With
\(\rho_m=0.35\), however, structural plasticity has a substantial context-dependent effect:
it can enable generalization in configurations without homeostasis, but generally delays
the attainment of perfect test accuracy when homeostasis is already active. Its contribution
to XOR is therefore not uniformly positive or negative, but depends strongly on the
interaction between connectivity sparsification and activity regulation.

\subsection{Mechanism-by-Mechanism Interpretation}

Homeostasis has the strongest and most consistent role across the two tasks. In sparse
parity, it regulates neuron use while structural plasticity forms a compact circuit. In XOR,
the original ablation with dense structural connectivity shows that homeostasis is sufficient
to produce generalization and is present in all successful configurations. The structural-density
ablation shows, however, that it is not strictly necessary: sufficiently sparse structural
plasticity can also support XOR generalization in its absence. Homeostasis is therefore best
interpreted as a robust regulator of hidden-neuron activity rather than as an indispensable
condition for XOR grokking.

Structural plasticity has a strong but strongly context-dependent role. In sparse parity,
where the target rule depends on only a small subset of the input coordinates, sparse
connectivity directly supports the formation of a compact effective circuit. In XOR, its
effect depends on structural density and on the presence of homeostasis. At
\(\rho_m=1\), the effective connectivity remains dense and structural plasticity has little
observable influence. At \(\rho_m=0.35\), structural sparsification can enable generalization
when homeostasis is absent, but it generally slows convergence to perfect test accuracy when
homeostasis is already active. Structural sparsification should therefore not be viewed as uniformly beneficial; its effect depends on whether sparse effective connectivity complements or competes with the other regulatory mechanisms.

Gain modulation is a secondary mechanism. Its contribution is limited in parity, but it complements homeostasis in XOR and appears in the best successful two-mechanism configuration. Its main role is consistent with response scaling: it changes the strength of neuron responses after the activity regime has been regulated.

Input gating also has a secondary role in these experiments. It is not required for successful generalization in either task, although it can support context-dependent routing of input information. Its contribution appears to depend more strongly on the structure of the task than the contributions of homeostasis or structural plasticity.

Lateral inhibition is also secondary in the present experiments. Removing it does not prevent generalization in either task, although XOR generalizes more slowly without it. However, this applies to the simplified implementation used here. Inhibitory pressure is computed from the mean activity of the other hidden neurons, without explicit neuron-to-neuron inhibitory connections or recurrent competitive dynamics. This choice keeps the computational cost reasonable. Biological lateral inhibition can involve direct and recurrent interactions among neurons and can support neural competition, response selectivity, and sparse, differentiated population codes. Homeostasis and structural plasticity may already provide part of the pressure toward neuron specialization in the present model. A richer implementation of lateral inhibition could therefore have a different effect.

Threshold modulation has a limited independent contribution in the present experiments. Its removal does not prevent generalization in either task, and the parity results can improve slightly without it. Its effect therefore appears to depend on interactions with the other regulatory mechanisms.

Activation decorrelation is also not required for successful grokking. Strong configurations can develop sparse and differentiated representations without the explicit decorrelation loss. The results therefore suggest that low redundancy can emerge from the other mechanisms of the model, even without a direct penalty on correlated hidden activity.

\subsection{Implications for Generalization and Large Language Models}

The results point to a broader view of generalization based on how a network organizes
and uses its internal capacity. Sparse parity and noisy XOR favor different forms of
organization. Parity benefits strongly from the formation of a compact effective circuit,
whereas XOR benefits from an internal representation that captures the latent cluster
structure. The structural-density results further show that such organization need not take
a unique form: different regulatory constraints can produce different representational
signatures while still supporting generalization. In both tasks, successful generalization
is therefore associated with a more structured allocation of computational capacity, but
the particular form of that structure depends on both the task and the mechanisms that
shape it.

Homeostasis provides the clearest common mechanism. Its main role is the regulation of neuron activity over training. It discourages persistent overuse or underuse of individual neurons and changes how computational capacity is distributed across the hidden layer. In parity, this regulation supports structural plasticity as the network concentrates the computation in a small circuit. In XOR, it supports stable specialization across latent regimes. The results therefore suggest that control of long-term component use can help a network develop representations that reflect the structure of the task.

This principle has an analogue in large language models (LLMs). A homeostatic mechanism could regulate the long-term activity or contribution of feed-forward neurons, channels, attention heads, experts, or other internal components. The regulated quantity would depend on the component. For an expert, it could be the fraction of tokens assigned to that expert. For a neuron or attention head, it could be a measure of activation or contribution to the layer output. Such regulation could discourage persistent concentration of computation in a small set of components while maintaining useful specialization. The exact threshold update used in the present model is only one possible implementation of this principle.

Structural plasticity suggests a second direction. In the present experiments, it changes
the effective circuit during training through persistent changes in the connectivity mask.
An LLM analogue could operate at a structured level and modify which feed-forward channels,
attention heads, experts, adapters, or blocks remain active in the effective model. Periodic
pruning, restoration, or reassignment of such components could allow the architecture to
reorganize as useful computational structure emerges. The parity results suggest that this
approach can be particularly effective when a task admits a compact or modular solution,
while the XOR density ablation shows that its benefit depends on the surrounding regulatory
regime. Structural sparsification can provide useful organization when other regulatory
pressures are weak or absent, but it can also slow convergence when combined with an
already effective mechanism such as homeostasis. This suggests that structural adaptation
should be treated as a context-dependent form of capacity allocation rather than as a
uniformly beneficial sparsity constraint.

Input gating suggests a different form of adaptation: context-dependent control of the information that enters a computation. Attention already provides input-dependent weighting of information, while mixture-of-experts architectures provide a more direct form of routing among computational components. A mechanism closer to the input gating studied here could modulate features or routing decisions according to the current context. The XOR results suggest a possible role for this type of control when different latent regimes benefit from different internal pathways.

Lateral inhibition provides a complementary principle of competition. In an LLM, such competition could operate among experts, attention heads, feature groups, memory components, or other sets of units that can develop overlapping roles. A competitive mechanism could suppress broad simultaneous recruitment and favor a clearer allocation of computation among specialized components. The present ablations place lateral inhibition below homeostasis and structural plasticity in overall importance, but its computational role remains relevant for architectures in which specialization and competition are central.

Activation decorrelation addresses redundancy more directly. An LLM analogue could penalize strongly correlated activity or encourage diversity among selected features, heads, or modules. The present experiments show that explicit decorrelation is not required for the emergence of differentiated representations, so it appears more suitable as a complementary objective than as a primary mechanism. Gain modulation and threshold modulation have a similarly secondary status in the ablation study. They provide additional forms of context-dependent control, but the present results give less reason to prioritize them over activity regulation, structural adaptation, routing, and competition.

The results suggest several directions for inclusion into LLMs. Homeostatic regulation could control the long-term use of computational components. Structural plasticity could reorganize the effective architecture during training. Input gating could provide context-dependent modulation or routing, and lateral inhibition could introduce competition among components with overlapping roles. These mechanisms address different aspects of the same problem: how an LLM allocates its capacity as useful internal structure develops. The experiments suggest that this allocation can be as important for generalization as the amount of available capacity itself.

\subsection{Computational Demands}

The proposed mechanisms differ substantially in computational cost. Their cost should be considered together with the number of optimization steps required to reach the desired level of generalization. A mechanism that adds computation to each training step can still be favorable if it substantially accelerates generalization. The ablation results show large differences in the time required to reach high test accuracy, so per-step overhead alone does not capture the full computational trade-off.

Homeostasis is one of the least expensive mechanisms in the model. For a minibatch of size $B$ and a hidden layer of width $n$, estimation of neuron activity requires a reduction over the $B\times n$ activation matrix, followed by an elementwise update of the $n$ threshold values. The additional computation therefore scales as $O(Bn)$, while the homeostatic state requires $O(n)$ memory. The threshold update is required only during training. At inference, the learned threshold values remain fixed and contribute only an elementwise operation to the forward pass.
Structural plasticity has a different computational profile. The model periodically updates a binary mask over the weight matrix by selecting connections according to weight magnitude. This introduces a global selection operation over the connections, but the update is performed only at specified intervals rather than at every optimization step. The computational effect at inference depends on how the final sparse structure is represented. A masked dense matrix retains most of the cost of dense computation. A sparse representation or a permanently pruned network can reduce computation, provided that the software and hardware can exploit the resulting sparsity.

Input gating, gain modulation, and threshold modulation require additional context-dependent projections. The model first computes the shared context vector and then uses separate projections to obtain the input gate, gain vector, and threshold shift. With a context dimension $s$, these operations add computation and parameters that scale with $s$ and the corresponding input or hidden dimensions. In the present model, $s=32$, so the modulation pathway remains relatively compact. The same projections remain active at inference because their outputs depend on the current input.

The implementation of lateral inhibition used here is computationally lightweight. For each sample, the inhibitory pressure can be computed from the sum of the hidden activities and an elementwise subtraction of the activity of each neuron. The additional cost therefore scales as $O(Bn)$. The mechanism introduces no neuron-to-neuron weight matrix and no recurrent iterations. This low cost is a direct consequence of the simplified global formulation used in the experiments.

Activation decorrelation has the least favorable scaling among the mechanisms considered here. For a minibatch activation matrix $H\in\mathbb{R}^{B\times n}$, the decorrelation loss constructs an $n\times n$ correlation matrix. The dominant matrix multiplication requires $O(Bn^2)$ computation, and storage of the correlation matrix requires $O(n^2)$ memory. This cost grows rapidly with hidden-layer width and makes explicit full-matrix decorrelation less attractive for very wide architectures unless an approximate or structured form is used.
Training and inference therefore have different computational profiles. Homeostatic updates, structural-mask updates, and the decorrelation loss are training-time operations. The final homeostatic thresholds and structural mask remain part of the trained model, while their update procedures are no longer required. Input gating, gain modulation, threshold modulation, and lateral inhibition remain part of the forward computation because they depend on the current input or hidden activity.

For LLMs, these differences favor mechanisms whose cost grows linearly with the number of active components or that produce structured sparsity that can be exploited efficiently. Homeostatic regulation and the simplified form of lateral inhibition have relatively small computational overhead. Structural plasticity can reduce inference cost when the learned structure can be represented efficiently. Context-dependent modulation introduces additional projections whose cost depends on the chosen modulation dimension. Explicit activation decorrelation is the most difficult mechanism to scale directly because of its quadratic dependence on representation width. These computational properties provide a practical basis for selecting which of the proposed mechanisms are most suitable for large-scale experiments.

\section{Conclusions}

This study shows that grokking can be influenced substantially by mechanisms that regulate how a network allocates and reorganizes its internal computational capacity. For both tasks, these mechanisms substantially accelerate the transition from memorization to generalization and can enable generalization in configurations that otherwise remain in a memorizing phase. The strongest effects come from homeostasis and structural plasticity. Homeostasis provides the most consistent improvement by regulating the long-term use of hidden neurons, while structural plasticity changes which connections participate in the effective computation. In sparse parity, these mechanisms work together to produce a highly compact and sparse effective circuit that generalizes rapidly. In noisy XOR, homeostasis provides a robust route to generalization, while the structural-density ablation shows that sufficiently sparse structural plasticity can itself induce generalization in configurations that otherwise fail. Successful generalization in noisy XOR is also accompanied by a pronounced reorganization of the hidden representation toward greater sparsity, specialization, and separation of the latent structure. Thus, grokking is facilitated when learning is accompanied by explicit pressure to reorganize the use of neurons and connections, rather than relying on weight optimization alone.

These results place internal reorganization alongside optimization-based explanations of grokking. After a model has learned to fit the training data, useful changes can still occur in how neurons are recruited, how computation is distributed, and which connections remain functionally important. Mechanisms that regulate these processes can therefore strongly accelerate the emergence of generalization and shape the internal organization that supports it.

Beyond the present experiments, the BioNN implementation provides a reusable testbed for evaluating the seven biologically inspired mechanisms on new tasks. Its independent switches preserve the same underlying network while allowing controlled comparisons of individual components and their combinations.

For LLMs, the question is whether similar regulation can shorten the phase in which internal representations continue to reorganize during training. Homeostatic control is attractive because its computational overhead is small, while structured plasticity offers a route toward adaptive sparse computation. If these principles promote earlier formation of stable and reusable internal structure at scale, they could reduce the optimization steps required for generalization. Biological mechanisms may thus provide useful design principles for improving learning efficiency.

%
%
%
%
%
%
%
%

\vskip 0.2in
\bibliography{sample}

\end{document}